\documentclass[10pt,twocolumn,letterpaper]{article}

\usepackage[pagenumbers]{iccv} 

\definecolor{iccvblue}{rgb}{0.21,0.49,0.74}
\usepackage[pagebackref,breaklinks,colorlinks,allcolors=iccvblue]{hyperref}
\usepackage{multirow}
\usepackage[T1]{fontenc}
\usepackage{algorithmic}
\usepackage{algorithm}
\usepackage{tabularx}
\usepackage{comment}
\usepackage[accsupp]{axessibility}

\hypersetup{
  pdftitle={DisCoPatch: Taming Adversarially-driven Batch Statistics for Improved Out-of-Distribution Detection},
  pdfauthor={Francisco Caetano, Christiaan Viviers, Luis A. Zavala-Mondragón, Peter H.N. De With, Fons van der Sommen}
}

\def\paperID{10217} 
\def\confName{ICCV}
\def\confYear{2025}

\title{DisCoPatch: Taming Adversarially-driven Batch Statistics for Improved Out-of-Distribution Detection}

\author{
Francisco Caetano$^{1,*}$, 
Christiaan Viviers$^{1,*}$, 
Luis A. Zavala-Mondragón$^{1}$,\\ 
Peter H.N. De~With$^{1}$, 
Fons van der Sommen$^{1}$ \\
$^{1}$Eindhoven University of Technology, The Netherlands \\
$^{*}$Equal contribution
}

\begin{document}
\maketitle
\begin{abstract}
Out-of-distribution~(OOD) detection holds significant importance across many applications. While semantic and domain-shift OOD problems are well-studied, this work focuses on covariate shifts - subtle variations in the data distribution that can degrade machine learning performance. We hypothesize that detecting these subtle shifts can improve our understanding of in-distribution boundaries, ultimately improving OOD detection. 
In adversarial discriminators trained with Batch Normalization~(BN), real and adversarial samples form distinct domains with unique batch statistics — a property we exploit for OOD detection. We introduce DisCoPatch, an unsupervised Adversarial Variational Autoencoder~(VAE) framework that harnesses this mechanism. During inference, batches consist of patches from the same image, ensuring a consistent data distribution that allows the model to rely on batch statistics. DisCoPatch uses the VAE's suboptimal outputs~(generated and reconstructed) as negative samples to train the discriminator, thereby improving its ability to delineate the boundary between in-distribution samples and covariate shifts. By tightening this boundary, DisCoPatch achieves state-of-the-art results in public OOD detection benchmarks. The proposed model not only excels in detecting covariate shifts, achieving 95.5\% AUROC on ImageNet-1K(-C), but also outperforms all prior methods on public Near-OOD~(95.0\%) benchmarks. With a compact model size of 25MB, it achieves high OOD detection performance at notably lower latency than existing methods, making it an efficient and practical solution for real-world OOD detection applications. The code is available at \href{https://github.com/caetas/DisCoPatch}{github.com/caetas/DisCoPatch}.
\end{abstract}    
\section{Introduction}
\label{sec:intro}

Out-of-distribution~(OOD) detection consists of identifying whether a given test sample significantly deviates from the known information of in-distribution~(ID) data. It is often employed as a preliminary step in image-based systems, aiming to mitigate the risks associated with feeding OOD inputs to a model. Besides safeguarding a system against erroneous predictions, it also facilitates the safe handling of OOD samples, either by rejection or transfer to human intervention. However, the significance of OOD lies not only in bolstering the reliability of image processing systems, but also in its standalone role for anomaly and fault detection. A simple example of this use case can be found in the visual inspection of industrial image data, where it is easy to acquire imagery of normal samples yet virtually impossible to define the expected defects~\cite{roth2022towards}. In the OOD context, these anomalies can be broadly classified into two types: (1)~anomalous objects in images which refer to unexpected or rare items appearing in the frame, and (2)~faulty equipment or products which refer to malfunctions or irregularities in the machinery or products under inspection. As a consequence, this task is typically cast as an OOD classification problem.

OOD detection comprises various types of shifts in data. (a)~Semantic shifts, such as encountering unseen classes, and (b)~domain shifts, like distinguishing between real images and drawings, have easily established boundaries and are well-defined in literature~\cite{hendrycks2016baseline,li2017deeper}. On the other hand, (c)~covariate shifts, which involve perturbations in data or subtle changes in its expected variability, are often conflated with domain shifts~\cite{yang2021generalized}. It is essential to differentiate covariate shifts, since they pose unique challenges requiring tailored detection mechanisms.

\begin{figure}[!t]
\centering
\includegraphics[width=\linewidth]{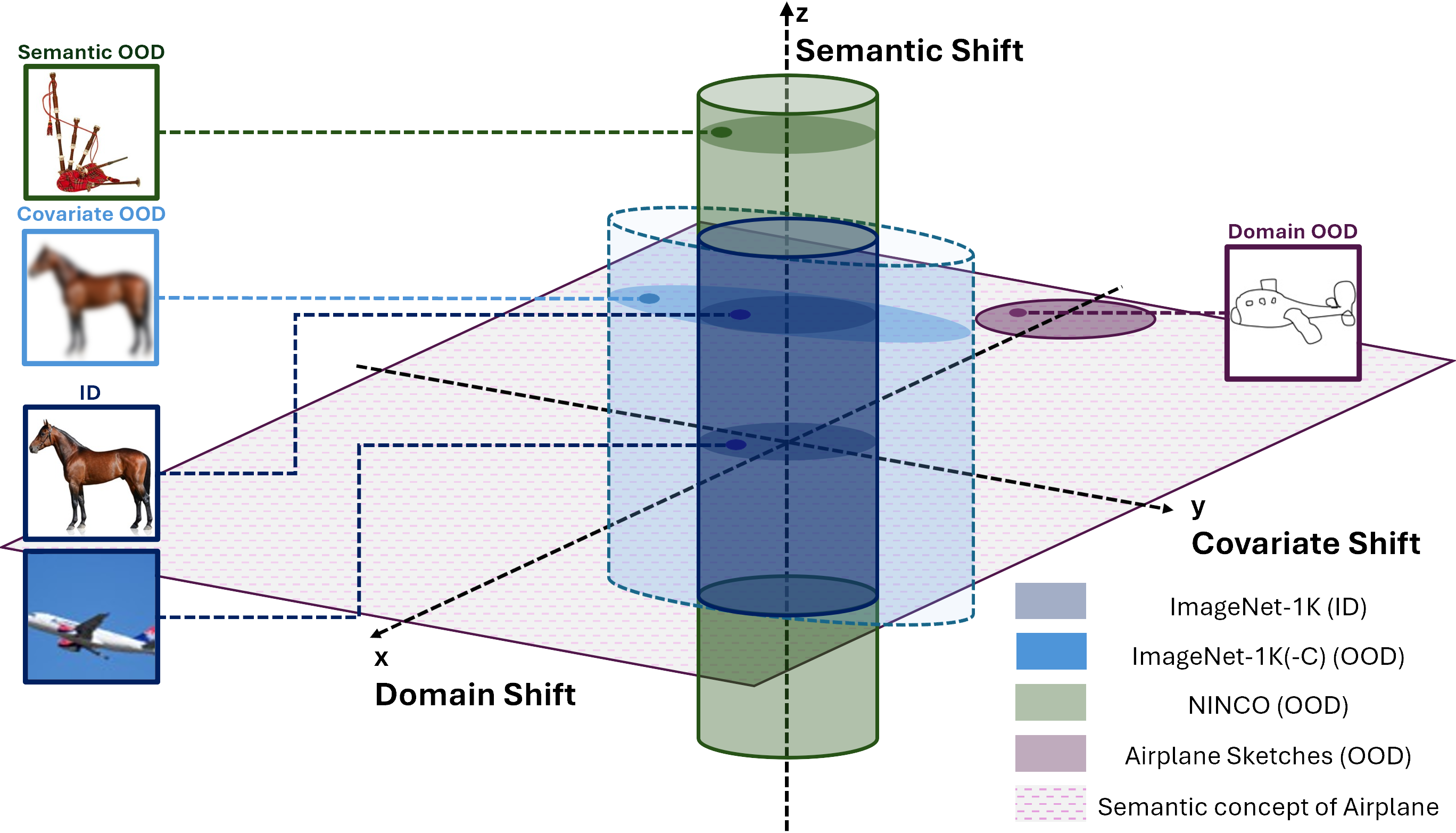}
\caption{Diagram illustrating data distribution shifts. Variations in the z-axis define semantic shifts, domain shifts represent new contexts like sketches with unchanged semantics and variability, and covariate shifts indicate changes within the same domain and semantic content, such as image perturbations (e.g., blurring).}
\label{fig:shifts}
\end{figure}

Figure~\ref{fig:shifts} illustrates the proposed framework for interpreting shifts in a data distribution. In our definition, the ID range covers an expected semantic shift, containing a pre-defined number of different classes, as exemplified by ImageNet-1K~\cite{russakovsky2015imagenet}, along with some degree of variability in terms of domain and covariate shifts. For instance, introducing a novel class such as bagpipes in NINCO~\cite{bitterwolf2023or} represents an OOD semantic shift, as ImageNet-1K lacks such examples. An extreme change in domain, such as a hand-drawn representation of a plane from the Sketch dataset~\cite{eitz2012hdhso}, is considered OOD, despite the retaining of semantic relevance. Additionally, substantial covariate shifts, such as a blurred horse image from ImageNet-1K(-C)~\cite{hendrycks2019robustness}, are also classified as OOD, even though there are no explicit alterations in semantic or domain concepts.

Various unsupervised OOD detection methods have been explored that utilize generative models, including Variational Autoencoders~(VAEs)~\cite{pinaya2021unsupervised}, Generative Adversarial Networks (GANs)~\cite{schlegl2017unsupervised}, Normalizing Flows~(NFs)~\cite{kobyzev2020normalizing} and more recently Denoising Diffusion Probabilistic Models~(DDPMs)~\cite{Wyatt_2022_CVPR}. The detection of anomalous data is usually performed by assessing whether they deviate from the learned representation manifold, or by comparing the reconstructed and original images in pixel space. DDPMs exhibit superior mode coverage compared to GANs and VAEs, albeit with much slower sampling/detection rates~\cite{xiao2021tackling}. NFs present a good framework for OOD detection, but it is well documented that they often assign a higher likelihood to OOD samples than the ID data~\cite{normalizingfail}. 

This paper proposes that the batch statistics extracted by Batch Normalization~(BN)~\cite{ioffe2015batch} layers can be used to improve OOD detection. It has been hypothesized that in adversarial networks trained with BN, clean and adversarial images are drawn from two distinct domains, with each domain exhibiting different means and variances in the BN layers~\cite{xie2019intriguing}. This has been empirically proved by examining feature statistics at different layers of a discriminator, thus validating that these stem from separate underlying distributions~\cite{wang2022removing}. This "two-domain hypothesis" points to BN’s inherent ability to separate ID and OOD samples based on batch statistics. However, BN increases adversarial vulnerability by shifting model reliance towards non-robust features rather than robust ones~\cite{benz2021batch}. To mitigate this, we propose utilizing a patch-based strategy for both training and inference. During training, partitioning images into patches encourages the model to focus on robust features that persist across diverse regions of the same image. During inference, processing batches of patches from a single image allows the model to ensure that the batch statistics correspond to the same underlying distribution.

In this paper, we demonstrate this effect can be leveraged by training a discriminator in an Adversarial VAE framework with DisCoPatch, using both reconstructed and generated images as OOD samples; this model yields an excellent OOD detector with an efficient and compact design. 
This approach not only excels in detecting covariate shifts but also proves effective against semantic OOD samples, all while significantly accelerating detection speed compared to prior methods. Since DisCoPatch is trained end-to-end, the quality of generated counterfactual samples becomes less critical; the VAE’s outputs naturally enhance the discriminator’s boundary-setting ability as training progresses. This method achieves high OOD detection performance with a compact and low-latency model, offering a practical solution for real-time applications. The main contributions of this work are as follows:
%
\begin{itemize} 
\setlength\itemsep{0em}
\item A novel analysis of Batch Normalization's inherent bias toward batch statistics, demonstrating how this mechanism can be leveraged for effective OOD detection by structuring batches with patches from the same image. 
\item DisCoPatch, a lightweight, unsupervised framework specifically designed for OOD detection.
\item State-of-the-art performance in Covariate Shift and Near-OOD detection, along with competitive results in Far-OOD detection, all while achieving significantly lower latency than existing methods.
\end{itemize}

\section{Related Work}

\subsection{Semantic Shift and Covariate Shift OOD}

OOD detection literature predominantly focuses on semantic shift and typically falls into two categories: (a)~supervised, which requires labels or OOD data, and (b)~unsupervised, which relies solely on ID data~\cite{yang2021generalized}. Given the nature of the OOD detection problem, OOD data are often not sufficiently representative, as OOD samples can come from a wide variety of unknown distributions. As such, unsupervised methods are generally preferred. 
Covariate shift occurs when images have consistent semantic and domain content, but are recorded under deviating imaging settings and conditions, or corrupted in a post-processing step. Although increasing the degree of variance under these conditions can deteriorate semantic and domain content, this study focuses on covariate shifts within the same domain, as these subtle distribution shifts can cause significant drops in the classification performance of machine learning models~\cite{hendrycks2019robustness}.

\subsection{Generative-based Methods}

A widely used and initially intuitive approach for OOD detection involves fitting a generative model $p(x; \theta)$ to a data distribution $x$ and evaluating the likelihood of unseen samples under this model, assuming that OOD samples will have lower likelihoods~\cite{bishop1994novelty}. However, this assumption has been challenged, with various generative models assigning higher likelihoods to certain OOD samples~\cite{hendrycks2018deep,nalisnick2018deep}. To address this, different approaches have been proposed, including using the Watanabe-Akaike Information Criterion (WAIC)~\cite{choi2018waic}, specific likelihood ratios~\cite{serra2019input,xiao2020likelihood}, and hierarchical VAEs~\cite{havtorn2021hierarchical}. These methods aim to correct for likelihood estimation errors, population-level background statistics, and model feature dominance. Another approach suggests labeling samples as OOD if their likelihoods fall outside the typical range of a model~\cite{chali2023improving,viviers2024can}, i.e., a sample may be classified as OOD not only if its likelihood is lower than that of ID data, but also if it is higher~\cite{morningstar2021density}.

\subsection{Reconstruction-based Methods}

Reconstruction-based methods involve training a model $R$ to reconstruct inputs $x$ from the training distribution, such that we obtain $\hat{x}=R(x)$. The rationale is that if $R$ has an information bottleneck, it will struggle to accurately reconstruct OOD inputs. However, these methods face practical challenges, including difficulty in tuning the information bottleneck size~\cite{pimentel2014review,denouden2018improving}. If it is too small, ID samples may not be faithfully reconstructed; if it is too large, the model can learn the identity function, allowing OOD samples to be reconstructed with low error. Some approaches address these issues by using the Mahalanobis distance in the Autoencoder's feature space as an OOD metric~\cite{denouden2018improving}, or by introducing a memory module to discourage OOD sample reconstruction~\cite{gong2019memorizing}. However, none of these methods fully resolve the bottleneck selection issue. To tackle this limitation, DDPMs have been employed, leveraging noise bottlenecks~\cite{Wyatt_2022_CVPR} and reconstructions from a range of noise values without the need for dataset-specific tuning~\cite{graham2023denoising} or of corrupted inputs~\cite{liu2023unsupervised}.

\subsection{Feature-based and Logit-based Methods} 

Several scoring functions have been devised to differentiate between ID and OOD examples, leveraging characteristics of ID samples, but not represented in OOD ones, and vice versa. These functions primarily stem from three sources: (1)~probability-based measures, such as maximum softmax probabilities~\cite{hendrycks2016baseline}, and minimum Kullback-Leibler~(KL) divergence between softmax and mean class-conditional distributions~\cite{hendrycks2019scaling}; (2)~logit-based functions, including maximum logits~\cite{hendrycks2019scaling}, and the use of the \texttt{logsumexp} function computed over logits~\cite{liu2020energy}; (3)~feature-based functions, involving the norm of the residual between a feature and its low-dimensional embeddings~\cite{ndiour2020out}, as well as minimum Mahalanobis distance between a feature and class centroids~\cite{lee2018simple}. Some hybrid methods combine both logit and feature scores for OOD detection~\cite{wang2022vim}, while more recent works have introduced masked image modeling pretraining into OOD detection with promising results~\cite{li2023rethinking,li2024moodv2}. However, the detection speed of these methods is severely constrained by their large transformer-based backbones.

\subsection{Adversarial Variational Autoencoders}

The VAE~\cite{kingma2013auto} consists of an encoder that predicts the parameters $\mu$ and $\sigma$ of the variational distribution of the input data, and a decoder that takes a sample from this distribution to reconstruct the input. VAEs are trained to maximize the Evidence Lower Bound (ELBO), which balances reconstruction fidelity with the latent space regularization to ensure that it follows a predefined probability distribution. Using latent space as a bottleneck restricts the information that can pass through, leading to uncertainty and blurriness in the reconstructions~\cite{dai2019diagnosing}. Additionally, the pixel-wise reconstruction error and the high dimensionality of natural image manifolds pose challenges for VAEs in generating high-quality and realistic samples. While natural images are assumed to lie on low-dimensional manifolds due to local scale redundancy~\cite{kretzmer1952statistics}, local details exist in higher-dimensional manifolds, making them difficult to capture.

GANs~\cite{goodfellow2014generative} consist of two neural networks with adversarial objectives: the generator learns to map a random vector to the data space; the discriminator acts as a classifier trained to differentiate real samples from generated ones. Despite their success in generation tasks, GANs suffer from two primary limitations compared to VAEs. The first is mode collapse, which occurs when the generator produces only a few different types repeatedly, making it easily recognizable by the discriminator. Consequently, the discriminator's feedback lacks useful information~\cite{thanh2020catastrophic}. Additionally, GANs lack an encoder network, which restricts their ability to reconstruct an input or manipulate its latent representation. AnoGAN~\cite{schlegl2017unsupervised} tries to circumvent this by optimizing a random latent vector to match a test sample and determining an anomaly score based on reconstruction quality and the discriminator’s output.

The VAE and GAN have been combined by incorporating a discriminator to enhance the realism of VAE reconstructions~\cite{larsen2016autoencoding}. Alternatively, the BiGAN~\cite{donahue2016adversarial} architecture features an encoder, generator, and discriminator, aiming for good unsupervised feature representations but tends to produce less accurate reconstructions. Other approaches have adapted this VAE/GAN combination to fully utilize the strengths of each architecture to improve the realism of the images produced by the model~\cite{plumerault2021avae}. DisCoPatch aims to retain the adversarial benefits and the mode coverage of the hybrid strategy, without the final goal of image generation, thereby reducing computational requirements.

\begin{figure*}[!t]
    \centering
    \includegraphics[width=\linewidth]{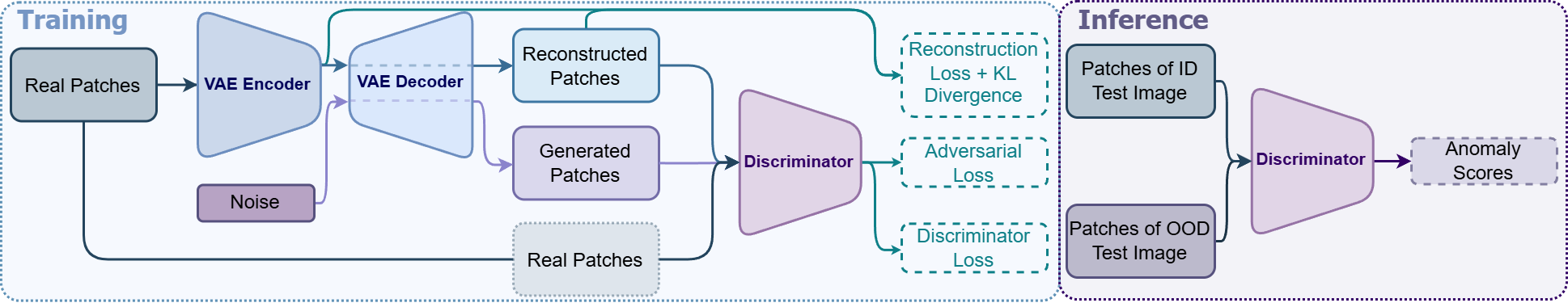}
    \caption{Overview of the DisCoPatch architecture. During inference, only the Discriminator is used.}
    \label{fig:DisCoPatch}
\end{figure*}

\subsection{Batch Normalization}
\label{subsec:batchnorm}

Batch Normalization~\cite{ioffe2015batch} is a widely used technique to speed up the training process of deep neural networks. It became popular for GAN architectures after being utilized in the DCGAN model~\cite{radford2015unsupervised} for both the generator and the discriminator. Essentially, BN takes a batch of feature samples \(\{x_1, x_2, \ldots, x_m\}\) and computes
\begin{equation}
 y_i = \frac{x_i - \mu_R}{\sigma_R} \cdot \gamma + \beta,   
\end{equation}
where, during training, \(\mu_R\) and \(\sigma_R\) are the running mean and the running standard deviation, which are updated according to the input batch statistics \(\mu_B\) and \(\sigma_B\) with a non-trainable \texttt{momentum} parameter $m$ by
\begin{equation}
\begin{bmatrix}
    \mu_R & \sigma_R
\end{bmatrix}
 = (1-m)
\begin{bmatrix}
    \mu_R & \sigma_R
\end{bmatrix}
+ m
\begin{bmatrix}
    \mu_B & \sigma_B
\end{bmatrix}\label{eq:BN}.
\end{equation}
Here, \(\gamma\) and \(\beta\) are learned parameters. It should be noted that when the BN layer is set to \emph{evaluation mode}, \(\mu_R\) and \(\sigma_R\) are fixed to the values learned throughout training~(i.e. parameter $m$ in Eq.~(\ref{eq:BN}) is set to zero).

Compared to models without normalization, BN accelerates training in the early stages and leads to better performance in GANs. However, the distinct distributions of clean and adversarial samples have hindered BN's effectiveness in adversarial settings~\cite{xie2019intriguing,wang2022removing}, as models trained with BN can still suffer from instability and low generalizability~\cite{xiang2017effects}. Alternative normalization approaches, such as Weight Normalization~\cite{salimans2016weight} and Spectral Normalization~\cite{miyato2018spectral}, have demonstrated improved performance for image generation and benefits in training stability, while still accelerating GAN training.

\section{DisCoPatch}
\label{sec:DisCoPatch}

\subsection{Overview}

A prevalent generative-based approach for OOD detection involves utilizing the trained generator to evaluate the likelihood of unseen samples. However, in adversarial setups, some information about the ID boundary will be incorporated into the discriminator, as it learns to assess the probability of a sample being real~(ID) or synthetic~(OOD). As mentioned in Section~\ref{sec:intro}, we exploit the observation that BN can help an adversarially trained discriminator to separate underlying data distributions by recognizing that clean and adversarial images are drawn from two distinct domains~(i.e. ID and OOD). Instead of solely estimating likelihoods, DisCoPatch clusters ID features while excluding OOD samples by employing an Adversarial VAE that generates and reconstructs samples to enhance feature separation. The discriminator penalizes poorly generated and reconstructed images, reinforcing the distinction between ID and OOD and refining the decision boundary.





It is on this premise that we propose a \textbf{Disc}riminative \textbf{Co}variate Shift \textbf{Patch}-based Network, DisCoPatch. DisCoPatch is an Adversarial VAE-inspired architecture, as shown in Figure~\ref{fig:DisCoPatch}, in which both the VAE and the discriminator are trained adversarially. Adversarially trained discriminators have been previously used for OOD detection~\cite{kong2021opengan}. DisCoPatch's approach combines generative and reconstruction-based strategies to distill information about the ID set and OOD boundaries to the discriminator during training in an unsupervised manner. Unlike traditional adversarial methods, DisCoPatch's focus is on leveraging the generator's output as a tool to refine the discriminator. DisCoPatch's discriminator only utilizes the current batch's (of patches) statistics in the \texttt{BatchNorm2D} layer~(i.e. parameter $m$ in Eq.~(\ref{eq:BN}) is set to one). Subsection~\ref{subsec:batch_methodology} details extensive ablation experiments to demonstrate its effectiveness.

\begin{figure}[!ht]
    \centering
    \includegraphics[width=\linewidth]{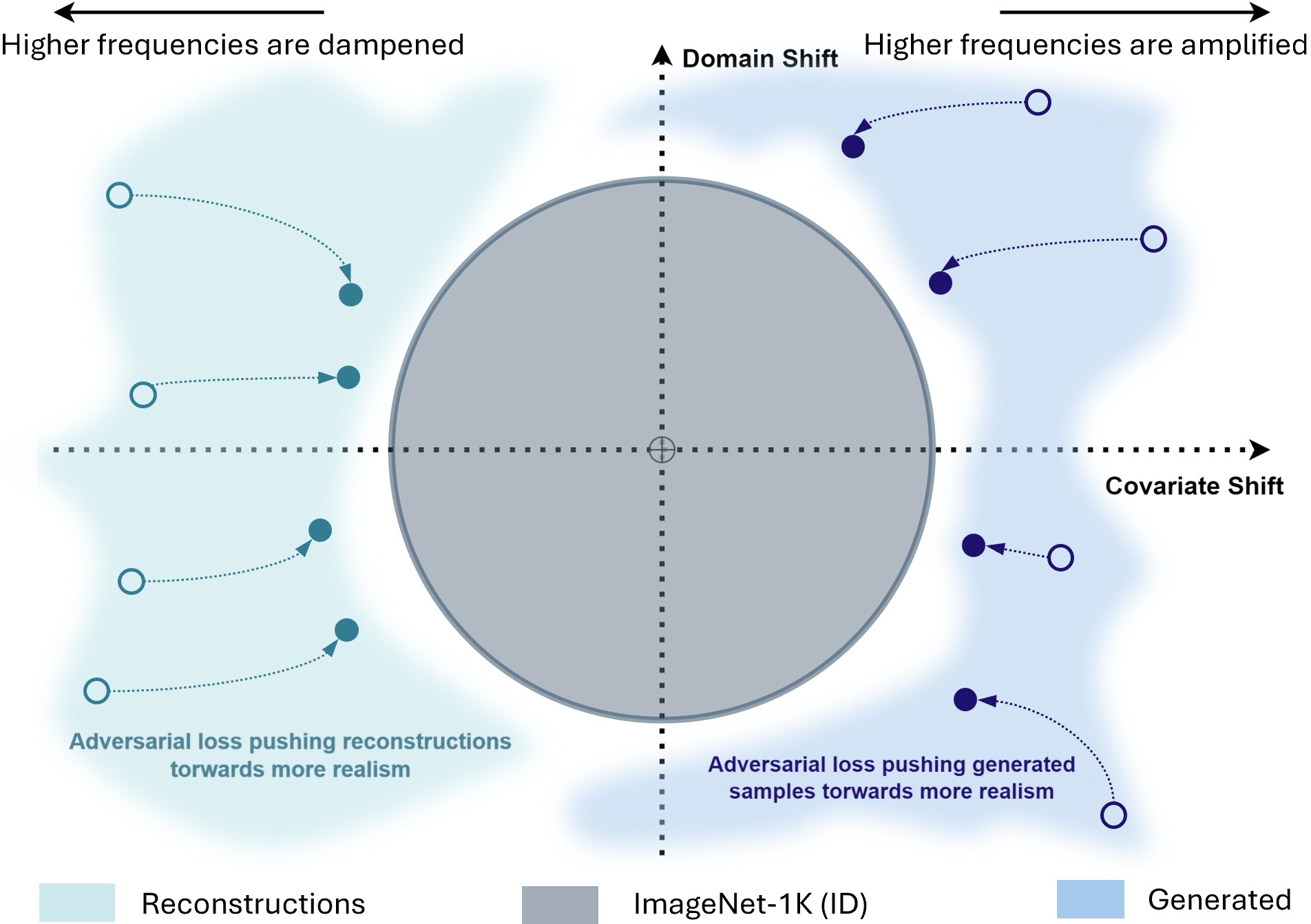}
    \caption{Covariate shifts can be simulated by reconstructed and generated patches. Encouraging more realism helps to tighten the border between the ID and the OOD sets.}
    \label{fig:frequency_shifts}
\end{figure}

\subsection{Training}
\label{subsec:training}

The VAE is trained to reduce the standard ELBO loss, while also producing samples (generated patches using the VAE decoder) that can fool the discriminator. The discriminator is trained to not only distinguish between generated and real patches, as in the standard GAN setup, but also reconstructed patches. Reconstructions from VAEs typically lack detail, i.e., they have a sub-optimal high-frequency representation~\cite{lin2023catch}, which can be found in certain types of covariate shifts, such as blurriness. On the other hand, images generated from GANs often exhibit severe high-frequency differences, leading the discriminator to focus excessively on these components~\cite{li2023exploring}. This focus can hinder the generator's ability to capture low-frequency components. By training the discriminator on reconstructions and generations, and encouraging both to appear more realistic, the discriminator's boundaries of the ID frequency spectrum become tighter, strengthening its ability to detect OOD samples, as illustrated in Figure~\ref{fig:frequency_shifts}.

The VAE in DisCoPatch's framework remains unchanged compared to the traditional VAE, with parameters $\theta$ and composed of an encoder $\mathcal{E}_{\theta_{\text{E}}}$ and a decoder $\mathcal{G}_{\theta_{\text{G}}}$ responsible for generating an image output. The VAE is a parameterized model given by $q_{\theta_{\text{E}}}(z|x^{(i)}) = \mathcal{N}(z; \mu^{(i)}, \sigma^{2(i)}\mathbf{I})$, where $\mu^{(i)}$ and $\sigma^{2(i)}$ are outputs of $\mathcal{E}_{\theta_{\text{E}}}$. The prior distribution of the latent codes is $p(z) = \mathcal{N}(z; 0, \mathbf{I})$. The VAE loss function combines a reconstruction term and a latent space regularization term, as demonstrated in the original paper by~\cite{kingma2013auto} and adversarial implementations~\cite{plumerault2021avae}. The reconstruction term optimizes the encoding-decoding process, while the regularization term aligns the encoder distributions with a standard Gaussian. The latter is represented by the KL-divergence between the predicted distribution and the prior distribution. Both terms are represented in Figure~\ref{fig:DisCoPatch} and can be written as
 \begin{equation}
\begin{split}
    \mathcal{L}_{\text{VAE}} &= \|x^{(i)} - \mathcal{G}_{\theta_{\text{G}}}(z)\|^2\\
    & -\frac{1}{2}\sum^{\text{dim}(z)}_{j=1} \left(1 + \log\left(\sigma^{2(i)}_j\right) - \mu^{2(i)}_j - \sigma^{2(i)}_j\right).
\end{split}
\end{equation}   
An additional model, the discriminator $\mathcal{D}$, parameterized by $\phi$, is added to the traditional VAE architecture. It has two main goals, as shown in Figure~\ref{fig:DisCoPatch}. First, it must discern between real patches and patches either reconstructed from $z_{\text{real}}$ or generated from random noise $z_{\text{fake}}$. This can be achieved by minimizing the cross-entropy function
\begin{equation}
\begin{split}
    \mathcal{L}_{\text{D}} & =  \mathbb{E}_{x\sim p(x)}\left[\log\left(1 - \mathcal{D}_{\phi}(x)\right) \right]\\
    & + \mathbb{E}_{x\sim p_{\theta_{\text{G}}}(x|z_{\text{real}})}\left[\log\left(\mathcal{D}_{\phi}(x)\right) \right] \\
    & + \mathbb{E}_{x\sim p_{\theta_{\text{G}}}(x|z_{\text{fake}})}\left[\log\left(\mathcal{D}_{\phi}(x)\right) \right].
\end{split}
\end{equation}
This suggests that in addition to the discriminator's initial goal of improving generated patches (sampled from random noise), it also pushes the reconstructions toward more realism. Therefore, an adversarial loss term, which encourages the VAE to generate or reconstruct patches that fool the discriminator, is added to the loss function, so that
\begin{equation}
\begin{split}
    \mathcal{L}_{\text{Adv}} & = \mathbb{E}_{x\sim p_{\theta_{\text{G}}}(x|z_{\text{real}})}\left[1 - \log\left(\mathcal{D}_{\phi}(x)\right) \right]\\
    & + \mathbb{E}_{x\sim p_{\theta_{\text{G}}}(x|z_{\text{fake}})}\left[1 - \log\left(\mathcal{D}_{\phi}(x)\right) \right].
\end{split}
\end{equation}
The final DisCoPatch loss function is thus a weighted combination of both the Vanilla VAE loss and the adversarial loss, which results in
\begin{equation}
\label{eq:vae_loss_final}
    \begin{split}
    \mathcal{L}_{\text{DCP}} & = \|x^{(i)} - \mathcal{G}_{\theta_{\text{G}}}(z)\|^2\\ 
    & -\frac{\omega_{\text{KL}}}{2}\sum^{\text{dim}(z)}_{j=1} \left(1 + \log\left(\sigma^{2(i)}_j\right) - \mu^{2(i)}_j - \sigma^{2(i)}_j\right) \\
    & +\omega_{\text{Rec}} \mathbb{E}_{x\sim p_{\theta_{\text{G}}}(x|z_{\text{real}})}\left[1 - \log\left(\mathcal{D}_{\phi}(x)\right) \right]\\
    & +\omega_{\text{Gen}} \mathbb{E}_{x\sim p_{\theta_{\text{G}}}(x|z_{\text{fake}})}\left[1 - \log\left(\mathcal{D}_{\phi}(x)\right) \right].
    \end{split}
\end{equation}
\subsection{Patching Strategy}

The patching strategy begins by taking a high-resolution input image, typically a standard 256$\times$256 resolution, and cropping it into \( N \) random patches, each of size 64$\times$64. This approach allows the model to capture fine-grained details in different image regions. During training, batches are composed of patches sourced from multiple images rather than from a single one. This setup accelerates training and ensures that the model learns consistent and robust ID features across a range of images, minimizing the potential for overfitting on individual image characteristics.

\begin{figure}[!t]
    \centering
    \includegraphics[width=\linewidth]{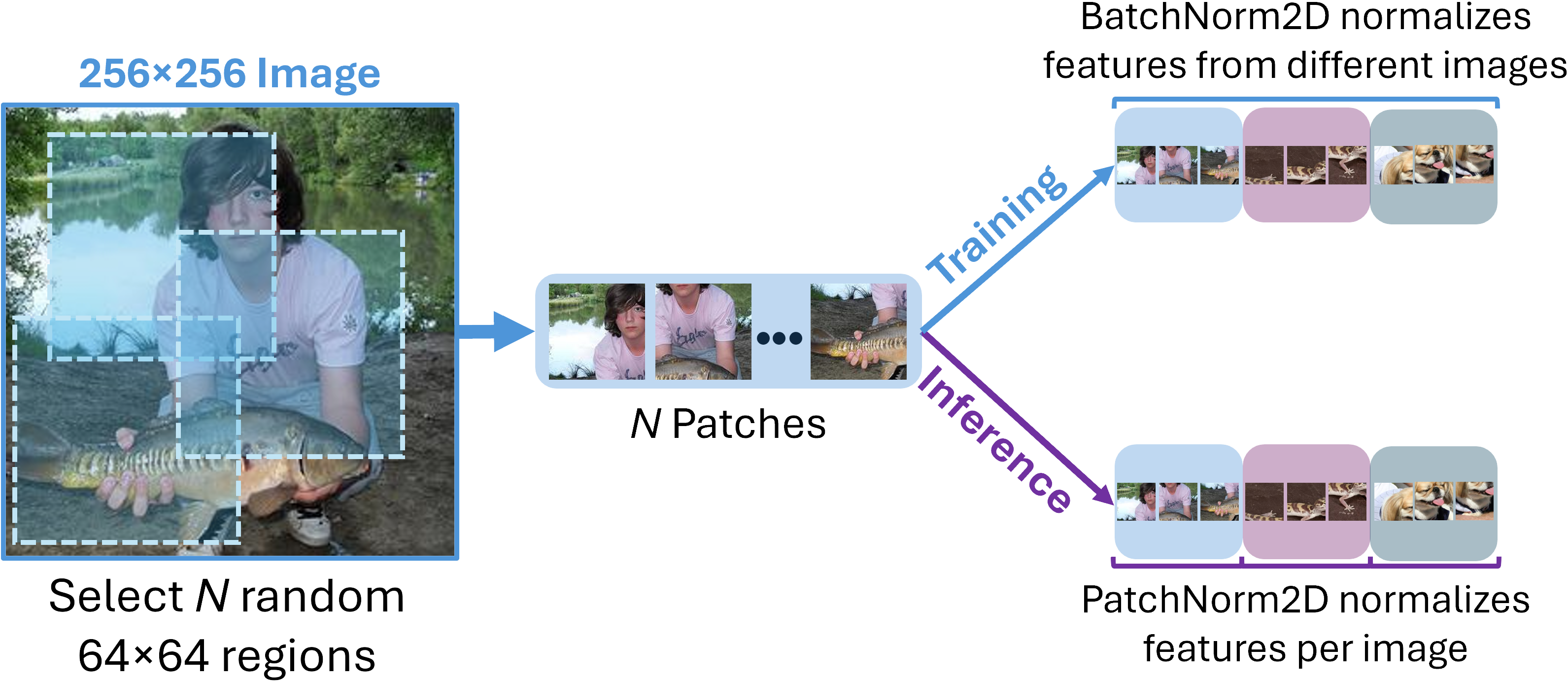}
    \caption{Different patching strategies employed by the model. Training normalizes patches from multiple images, while inference normalizes patches per image.}
    \label{fig:patches}
\end{figure}

In this setup, using the standard \texttt{BatchNorm2D} layer constrains batch processing during inference to a single image, since normalization must be applied only to features from the same distribution; mixing ID and OOD samples in a batch would distort the statistics and hinder effective separation. Therefore, we introduce a custom layer, \texttt{PatchNorm2D}, which retains the weights and biases of \texttt{BatchNorm2D} but enables the simultaneous normalization of multiple groups of patches. Specifically, each batch consists of several groups of $N$ patches, where each group contains patches from the same image that are normalized together. This ensures that the model produces \textit{independent} results per image while maintaining the benefits of batch normalization. The final anomaly score for an image is the mean of the scores of all patches within that same image. For the remainder of the paper, we refer to the model as DisCoPatch-$N$, indicating the number of patches per image used during inference. Both strategies are illustrated in Figure~\ref{fig:patches}.

%
\section{Experiments \& Methodology}

\subsection{Datasets}
In OOD detection benchmarks, the conventional approach involves designating an entire dataset as ID and then compiling multiple datasets that lack any semantic overlap with the ID categories to act as OOD sets.
%
To ensure consistency in the benchmarking process, we adhere to the methodology proposed by OpenOOD~\cite{yang2022openood}.
Our evaluation encompasses three tasks: (1)~\textit{Near-OOD}, which exhibits slight semantic variation compared to ID datasets; (2)~\textit{Far-OOD}, which encompasses both semantic and domain shifts; and (3)~\textit{Covariate Shift OOD}, involving corruptions within the ID set. ImageNet-1K~\cite{russakovsky2015imagenet} was defined as the ID dataset. Further details on the datasets are summarized in Appendix~\ref{subsec:data_sup}.

\subsection{Evaluation Metrics}

The evaluation metrics employed in OpenOOD by~\cite{yang2022openood} are adopted for this work. These two main evaluation metrics are: (1)~\textit{AUROC}, which measures the area under the Receiver Operating Characteristic (ROC) curve, and displays the relationship between True Positive Rate (TPR) and False Positive Rate (FPR); and (2)~\textit{FPR95}, which measures the FPR when the TPR is equal to 95\%, with lower scores indicating better performance. The full results are provided in the form "AUROC/FPR95\%". In addition to these quantitative metrics, a UMAP visualization of the extracted features is included as a qualitative evaluation, providing insight into the separability of ID and OOD samples in the learned feature space.
\subsection{Baseline Models}
For the ImageNet-1K benchmark, we compare our method, DisCoPatch, against SOTA public models, such as \textit{MOODv2}~\cite{li2024moodv2}, \textit{NNGuide}~\cite{park2023nearest}, \textit{SCALE}~\cite{xu2023scaling}, RankFeat\&Weight~\cite{song2024rankfeat_short}, ASH~\cite{djurisic2022extremely_short}, and FDBD~\cite{liu2023fast_short}, all of which have demonstrated SOTA performance\footnote{OpenOOD Benchmark \href{https://zjysteven.github.io/OpenOOD/}{https://zjysteven.github.io/OpenOOD/}} on Near-OOD and Far-OOD detection for these datasets. These models employ feature-based and logit-based strategies to perform OOD detection. In the case of NNGuide, we evaluate three of its available backbones to assess performance across different model types: the top-performing RegNet~\cite{radosavovic2020designing}, a more efficient ResNet-50~\cite{he2016deep}, and the lightweight MobileNetV2~\cite{sandler2018mobilenetv2}.

\subsection{DisCoPatch Implementation \& Compute}

Appendix~\ref{subsec:implementation_sup} contains the implementation specifics of DisCoPatch. The computational resources for training and evaluating models are detailed in Appendix~\ref{subsec:compute_sup}.

\subsection{Experimental Significance}

To support the main claims of this work, we perform five random runs to validate DisCoPatch's performance on the OOD benchmarks and compute the average performance.

\subsection{Batch Normalization Bias Analysis}
\label{subsec:batch_methodology}

We conduct various supplemental experiments to investigate the effect of Batch Normalization’s reliance on the batch statistics $\mu_{B}$ and $\sigma_{B}$ (Eq.~\ref{eq:BN}) during training. We conduct the experiments using the complete ImageNet-1K dataset, center-cropped and resized to \(256\times256\) pixels, within the DisCoPatch framework. Unlike our main experiments that rely on patch-based batches, here we evaluate the model's sensitivity to batch-level statistics across the full-resolution images. This model was used with the standard \texttt{BatchNorm2D} parameters, with \texttt{track\_running\_stats} set to \texttt{True} and the default \texttt{momentum} of 0.1. 

In the first experiment, the model is analyzed in \textit{evaluation mode}, as described in Subsection~\ref{subsec:batchnorm}. In this mode, the model utilizes the running mean, $\mu_{R}$, and variance, $\sigma_{R}$, learned during training and neglects the statistics of the current batch. As mentioned in Section~\ref{sec:intro}, this paper hypothesizes that the batch statistics of the batch being evaluated can be a powerful feature for OOD detection. Consequently, we also introduce an additional model, in which the \texttt{track\_running\_stats} option in PyTorch’s \texttt{BatchNorm2D} layer is set to \texttt{False}. This setting causes the model to disregard the running mean and variance obtained during training, and instead, it employs solely the statistics $\mu_{B}$ and $\sigma_{B}$ of the batch being tested for normalization. This configuration does not modify the layer weights and is solely used to illustrate dependency on batch-level statistics during inference. For a comprehensive analysis, refer to Appendix~\ref{subsec:batch_norm_bias}. Alternative normalization strategies are covered in Appendix~\ref{subsec:alt_sup}.
\begin{table*}[!t]
    \centering
    \caption{OOD benchmark results for models trained on ImageNet-1K. Results are obtained from the respective research papers where available or recomputed~($^{*}$) and reported following the OpenOODv1.5 benchmark as AUROC/FPR95.}
    \label{tab:in-1k-table}
    {\resizebox{1.0\linewidth}{!}{
    \begin{tabular}{@{\extracolsep{10pt}}lcccccc@{}}
        \toprule
         \multirow{2}{*}{\textbf{Model}} & \multicolumn{2}{c}{\textbf{Near-OOD}} & \multicolumn{3}{c}{\textbf{Far-OOD}} & \textbf{Covariate Shift}\\
         \cline{2-3}\cline{4-6}\cline{7-7}
         & \textbf{SSB-hard} & \textbf{NINCO} & \textbf{iNaturalist} & \textbf{DTD} & \textbf{OpenImage-O} & \textbf{ImageNet-1K(-C)}\\
         \midrule
         MOODv2~\citep{li2024moodv2}~(BEiTv2) & \phantom{$^{*}$}85.0/58.1$^{*}$ & \phantom{$^{*}$}92.7/38.2$^{*}$ & 99.6/1.8 & 94.3/24.7 & 97.4/13.6 & \phantom{$^{*}$}70.5/73.9$^{*}$\\
         SCALE~\citep{xu2023scaling}~(ResNet-50) & 77.4/67.7 & 85.4/51.8 & 98.0/9.5 & \textbf{97.6/11.9} &  94.0/28.2 & \phantom{$^{*}$}83.3/54.1$^{*}$ \\
         NNGuide~\cite{park2023nearest}~(RegNet) & \phantom{$^{*}$}84.7/54.7$^{*}$ & \phantom{$^{*}$}93.7/\textbf{28.9}$^{*}$ & \textbf{99.9/1.8} & 95.8/17.0 & \textbf{97.7/10.8} & \phantom{$^{*}$}78.5/61.6$^{*}$ \\
         NNGuide~\cite{park2023nearest}~(ResNet-50) & \phantom{$^{*}$}71.5/82.8$^{*}$ & \phantom{$^{*}$}80.5/69.8$^{*}$ & 96.9/14.3 & 90.4/27.4 & 92.4/35.4 & \phantom{$^{*}$}78.6/58.7$^{*}$ \\
         NNGuide~\cite{park2023nearest}~(MobileNetV2) & \phantom{$^{*}$}64.3/90.3$^{*}$ & \phantom{$^{*}$}75.5/80.2$^{*}$ & 80.2/70.2 & 88.3/40.6 & 82.9/63.3 & \phantom{$^{*}$}78.1/61.4$^{*}$ \\
         RankFeat~\cite{song2024rankfeat}~(ResNetv2-101) & \phantom{$^{*}$}89.4/47.9$^{*}$ & \phantom{$^{*}$}90.0/39.3$^{*}$ & 96.0/13.0 & 95.0/25.4 & \phantom{$^{*}$}92.4/33.3$^{*}$ & \phantom{$^{*}$}91.3/38.7$^{*}$ \\
         ASH~\cite{djurisic2022extremely}~(ResNet-50) & \phantom{$^{*}$}74.1/80.6$^{*}$ & \phantom{$^{*}$}83.0/64.2$^{*}$ & 97.9/11.5 & \textbf{97.6/11.9} & \phantom{$^{*}$}92.8/32.7$^{*}$ & \phantom{$^{*}$}84.7/51.3$^{*}$\\
         FDBD~\cite{liu2023fast}~(ResNet-50) & \phantom{$^{*}$}68.4/83.4$^{*}$ & \phantom{$^{*}$}81.4/66.1$^{*}$ & 97.6/12.4 & 98.0/10.7 & \phantom{$^{*}$}91.7/35.6$^{*}$ & \phantom{$^{*}$}82.2/57.8$^{*}$\\
         DisCoPatch-64~(Proposed) & \textbf{95.8/19.8} & \textbf{94.3}/39.0 & 99.1/3.6 & 96.4/18.9 & 94.4/29.7 & \textbf{97.2/10.6} \\
         \bottomrule
    \end{tabular}}}
\end{table*}

\section{Results}
\label{sec:experiments}



Table~\ref{tab:in-1k-table} shows that DisCoPatch-64 surpasses state-of-the-art methods in Near-OOD and Covariate Shift benchmarks, with a particularly large performance gap in the Covariate Shift OOD detection task. Although DisCoPatch does not achieve SOTA performance in any of the Far-OOD benchmarks, it is close to matching the best performers on the iNaturalist and DTD tasks and attains competitive performance in OpenImage-O, whilst being a much smaller model. Detailed performance analysis for Covariate Shift can be found in Appendix~\ref{subsec:results_sup_in1k}.

\begin{figure}[!ht]
    \centering
    \includegraphics[width=\linewidth]{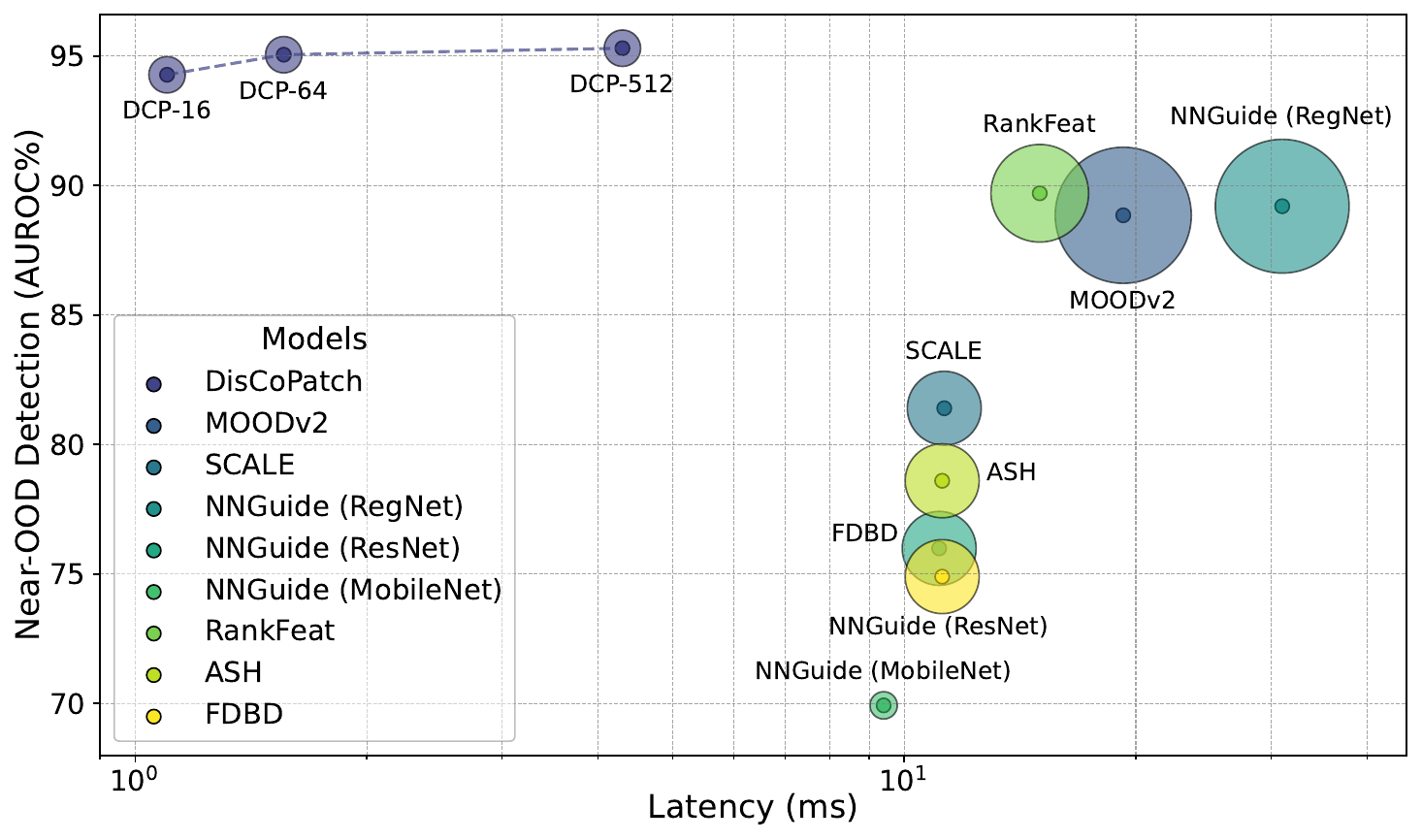}
    \caption{AUROC on Near-OOD detection vs. latency of the models. Circumference size is equivalent to relative model size.}
    \label{fig:nearood}
\end{figure}

The superiority of DisCoPatch extends beyond raw detection performance. As illustrated in Figure~\ref{fig:nearood}, DisCoPatch outperforms all other models in Near-OOD detection performance while reducing latency by up to an order of magnitude. Even with only 16 patches per image, DisCoPatch outperforms all other models tested. Appendix~\ref{subsec:count_sup} offers a detailed examination of the performance of additional patch counts. However, increasing the patch count beyond 64 did not result in any noticeable performance improvements. The performance gap between DisCoPatch and the second fastest model, NNGuide (MobileNetV2), is near 25\%.

\begin{figure}[!t]
    \centering
    \includegraphics[width=\linewidth]{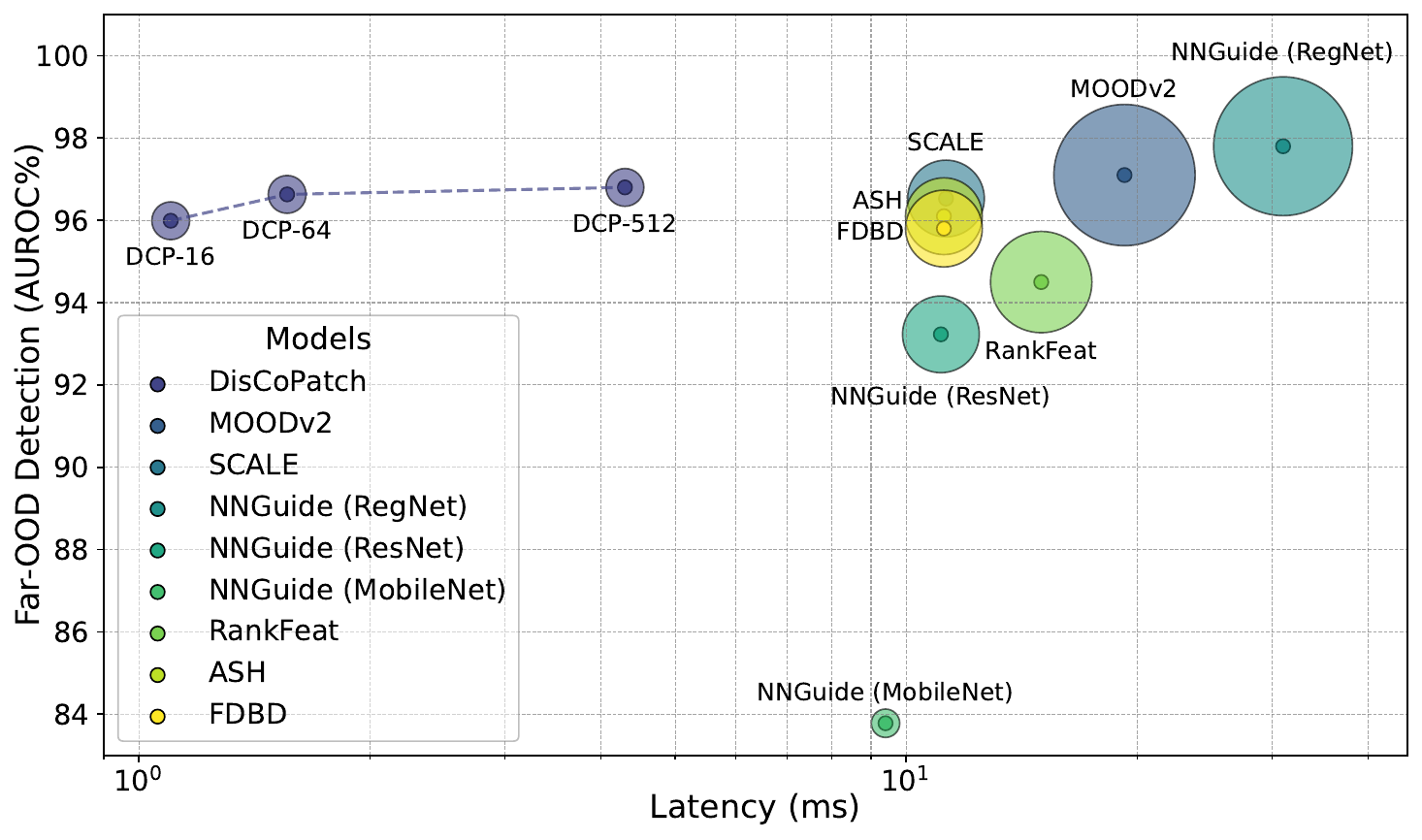}
    \caption{AUROC on Far-OOD detection vs. latency of the models. Circumference size is equivalent to relative model size.}
    \label{fig:farood}
\end{figure}


Although DisCoPatch does not achieve the highest performance in Far-OOD detection, Figure~\ref{fig:farood} shows that it is surpassed only by the two largest and slowest models, MOODv2 and NNGuide with a RegNet backbone, with a marginal gap of just 1\%, while being at least ten times faster. DisCoPatch also exceeds the performance of NNGuide (MobileNetV2), the next smallest model, by 12\%. Again, increasing the number of patches per image beyond 64 yielded no further significant improvements in detection performance.
Table~\ref{tab:inference} in the supplementary material provides details on model size and latency.

\begin{figure}[!t]
    \centering
    \includegraphics[width=\linewidth]{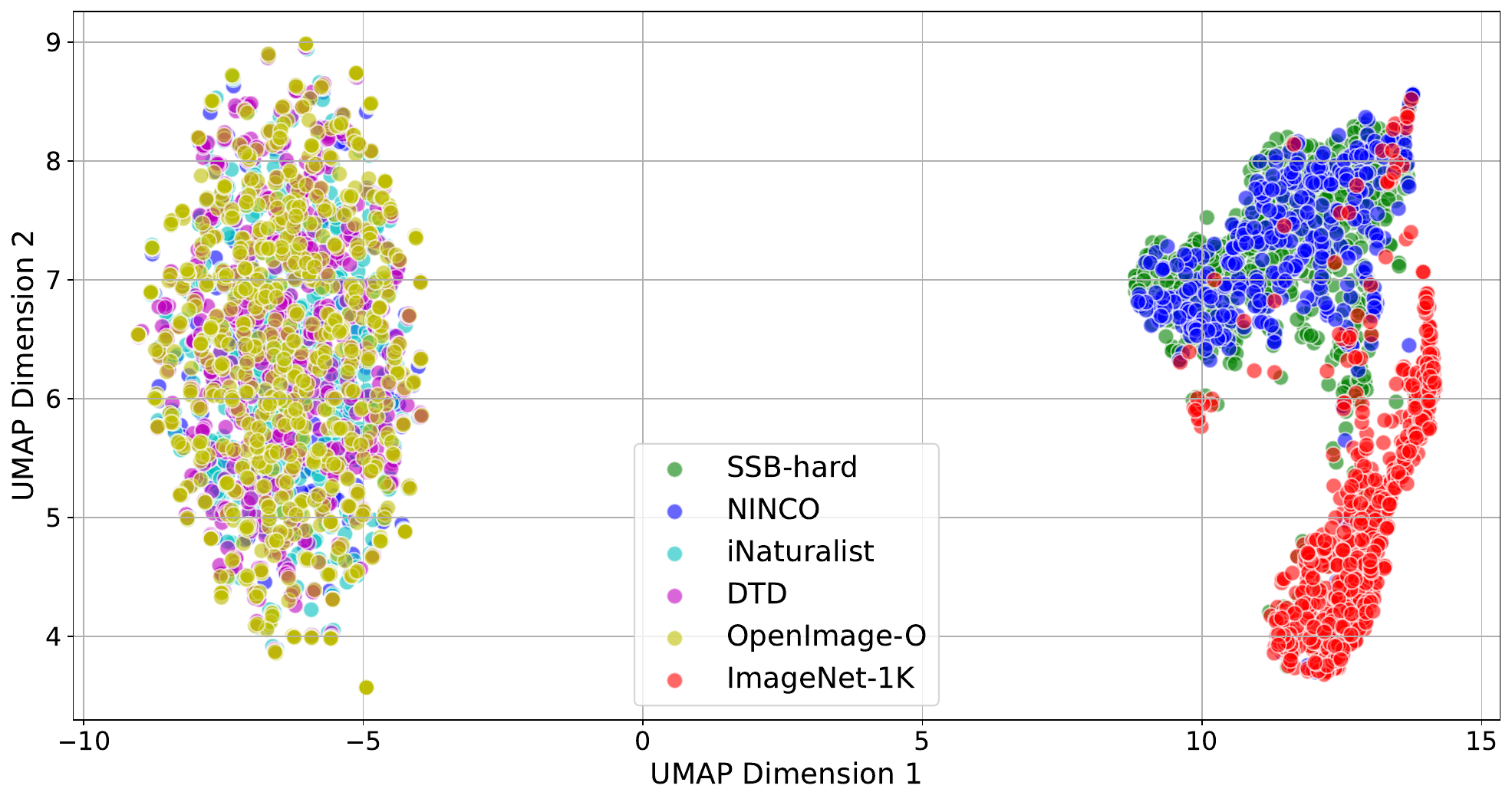}
    \caption{UMAP visualization of discriminator-extracted features, each point corresponding to the mean features of image patches.}
    \label{fig:umap}
\end{figure}

Figure~\ref{fig:umap} presents a UMAP visualization of the average features extracted by the discriminator from 64 patches per image. The results show that samples from Near-OOD datasets~(NINCO, SSB-hard) cluster closer to the ID set~(ImageNet-1K), while Far-OOD samples~(OpenImage-O, DTD, iNaturalist) are mapped farther away. While the proposed method effectively structures the feature space into distinct clusters, the decision boundary remains suboptimal. This highlights the potential for integrating post-hoc feature-based methods to further enhance OOD separation.

\section{Discussion}


\textit{Covariate Shift OOD.}~Validation on ImageNet-1K(-C) reveals that DisCoPatch achieves a substantial improvement over other models in this task. As hypothesized in Section~\ref{sec:DisCoPatch}, and supported by recent findings~\citep{lin2023catch,li2023exploring}, training a discriminator with VAE reconstructions enhances sensitivity to corruptions that diminish the high-frequency spectrum. This effect arises because reconstructed images generally lack high-frequency content, prompting the discriminator to classify this lack of content as "fake" and its presence as "real." In contrast, training the discriminator with generated images strengthens its ability to detect high-frequency amplification. DisCoPatch’s unique unsupervised training approach, which exposes the discriminator to generated and reconstructed patches, enables robust detection of low- and high-frequency perturbations, leading to consistent performance across various corruptions.

\textit{Near-OOD and Far-OOD.}~The covariate shift-focused training strategy effectively tightens the boundary between ID and OOD samples, improving detection performance on both Near-OOD and Far-OOD datasets. DisCoPatch outperforms existing models, achieving SOTA performance in Near-OOD detection. Although the same does not occur for Far-OOD, DisCoPatch's performance is competitive, and the model is only beaten by far larger and slower models.

\textit{Deployment.}~There are generally two main deployment scenarios for OOD detection algorithms: (1)~The OOD detection algorithm is the primary focus, deployed as a standalone application. (2)~The OOD detection algorithm operates alongside a main image processing algorithm, ensuring its safe and effective use. An OOD algorithm must be practical and effective in real-world scenarios, delivering strong detection performance while being highly deployable. Deployability should be assessed in the following aspects: (a)~\textit{Accessibility}, evaluated by the compute requirements necessary for the algorithm; (b)~\textit{Development Cycle}, measured by the time required for model training and deployment; (c)~\textit{Inference Speed}, which reflects the time it takes for the algorithm to make predictions during deployment; and (d)~\textit{Accuracy}, denoting the ability of the algorithm to provide highly accurate OOD detection.

An ideal OOD detection algorithm excels in all these dimensions, ensuring that it can be used effectively in various practical applications. As detailed in our results and Appendix~\ref{subsec:compute_sup}, DisCoPatch excels in all criteria. The model achieves SOTA OOD detection results, while utilizing a substantially smaller and faster model. Table~\ref{tab:inference} indicates that DisCoPatch is 1 order of magnitude faster than the other evaluated models; being up to 12 times faster than MOODv2 and up to 19 times quicker than NNGuide. Training a tailored model is also efficient and flexible, as demonstrated in Table~\ref{tab:comp}, making it a strong candidate for applications that require fast development cycles.
\section{Limitations \& Future Work}
\label{sec:limitations}
Comprehensive benchmarking of Covariate Shift detection across diverse architectures is vital to advance the field. While DisCoPatch shows promise for generative-based setups, evaluating more models will clarify how architecture impacts detection performance. Covariate Shift detection is particularly critical in fields like medical imaging, where identifying distribution shifts is crucial to ensuring trustworthy predictions. Expanding such evaluations to high-stakes domains could broaden the relevance of DisCoPatch and similar models for robust, real-world applications.
While current results demonstrate the effective use of batch statistics for OOD detection, a signal-processing-focused analysis of feature propagation and suppression at each layer could provide key insights, improving model interpretability. The planned extensions include the integration of several ResNet-based architectures that use \texttt{BatchNorm2D} layers. Beyond discriminator architectures, future work should also assess more robust reconstructors (e.g., VQ-GAN) and generators (e.g., DDPMs). 

\section{Conclusion}
\label{sec:conclusions}

This paper introduces DisCoPatch, an unsupervised lightweight framework for OOD detection. DisCoPatch, unlike traditional VAE or GAN training objectives, uses a combination of reconstructed and generated images to address a wide range of frequency-spectrum perturbations. Furthermore, this study highlights the inherent bias in Batch Normalization toward batch statistics. DisCoPatch effectively exploits this bias for competitive OOD detection, particularly for Covariate Shift and Near-OOD detection. DisCoPatch sets a new state-of-the-art in public OOD detection benchmarks, with 95.5\% AUROC on Covariate Shift detection and 95.0\% on Near-OOD detection, while maintaining low latency and a compact 25MB model, which makes it well suited for real-time, resource-constrained applications. 

\section*{Acknowledgments}

This work used the Dutch national e-infrastructure with the support of the SURF Cooperative using grant no. EINF-10021. This research was funded by the European Xecs Eureka TASTI Project.
{
    \small
    \bibliographystyle{ieeenat_fullname}

}

\definecolor{darkgreen}{rgb}{0,0.4,0}
\clearpage
\setcounter{page}{1}
\maketitlesupplementary

\appendix

The supplementary material is organized as follows: Appendix~\ref{subsec:data_sup} covers the datasets used in this work. Appendix~\ref{subsec:implementation_sup} describes the implementation details of DisCoPatch while Appendix~\ref{subsec:compute_sup} covers the compute resources required for training and evaluating the models. Appendix~\ref{subsec:batch_norm_bias} presents the results of the experiments on Batch normalization bias in conventional setups. Appendix~\ref{subsec:alt_sup} covers the impact of alternative normalization layers in the discriminator performance. Appendix~\ref{subsec:count_sup} analyses in more detail the effect that multiple patch counts have on DisCoPatch's performance. Appendix~\ref{subsec:results_sup_in1k} provides detailed results on the ImageNet-1K Covariate Shift OOD detection benchmark.

\section{Data Availability}
\label{subsec:data_sup}

\textit{ImageNet-1K}~\citep{russakovsky2015imagenet} contains 1000 classes. For the Near-OOD experiments in this paper, we have employed the SSB-hard~\citep{vaze2021open} and NINCO~\citep{bitterwolf2023or} datasets. In the case of Far-OOD, we have used iNaturalist~\citep{huang2021mos}, DTD~\citep{cimpoi14describing}, and OpenImage-O~\citep{wang2022vim}. The experiments in the Covariate Shift section are evaluated in the ImageNet-1K(-C)~\citep{hendrycks2019robustness} dataset. The OOD benchmark used to evaluate and compare the selected models closely follows the one proposed in OpenOOD by~\cite{yang2022openood}. The images are resized to 256$\times$256 before being fed to DisCoPatch.

The dataset ImageNet-1K(-C) was downloaded from its source~\footnote{\href{https://github.com/hendrycks/robustness}{https://github.com/hendrycks/robustness}}. Additionally, we have also used the original and publicly available splits for ImageNet-1K~\footnote{\href{https://huggingface.co/datasets/benjamin-paine/imagenet-1k-256x256}{https://huggingface.co/datasets/benjamin-paine/imagenet-1k-256x256}}. The remaining datasets and files containing training and evaluation splits were downloaded from OpenOOD's publicly available repository~\footnote{\href{https://github.com/jingkang50/openood}{https://github.com/jingkang50/openood}}. For convenience, DisCoPatch's repository includes the split files and a script that automatically downloads these datasets. 

\section{Implementation Details}
\label{subsec:implementation_sup}

\begin{algorithm}[!ht]
\caption{Training algorithm of DisCoPatch.}
\label{alg:train}
\begin{algorithmic}
\STATE 
\STATE {\text{Initialize parameters of models }$\theta, \phi$}
\STATE \hspace{0.3cm}$ \textbf{while } \text{training} \textbf{ do} $
\STATE \hspace{0.6cm}$ x^{real} \gets \text{patches of images from dataset} $
\STATE \hspace{0.6cm}$ z^{real}_{\mu},z^{real}_{\sigma} \gets \mathcal{E}_{\theta_{\text{E}}}(x^{real})$
\STATE \hspace{0.6cm}$ z^{real} \gets z^{real}_{\mu} + \epsilon_{real} z^{real}_{\sigma} \text{ with } \epsilon_{real} \sim \mathcal{N}(0,\mathbf{I})$
\STATE \hspace{0.6cm}$ x^{rec} \gets \mathcal{G}_{\theta_{\text{G}}}(z^{real})$
\STATE \hspace{0.6cm}$ z^{fake} \gets \epsilon_{fake} \text{ with } \epsilon_{fake} \sim \mathcal{N}(0,\mathbf{I})$
\STATE \hspace{0.6cm}$ x^{fake} \gets \mathcal{G}_{\theta_{\text{G}}}(z^{fake})$
\STATE \hspace{0.6cm}$ x^{rec} \gets \mathcal{G}_{\theta_{\text{G}}}(z^{real})$
\STATE \hspace{0.6cm}$ D^{real} \gets \mathcal{D}_{\phi}(x^{real})$
\STATE \hspace{0.6cm}$ D^{rec}, D^{fake} \gets \mathcal{D}_{\phi}(x^{rec}),\mathcal{D}_{\phi}(x^{fake})$
\STATE
\STATE \hspace{0.6cm}$ \theta \bar{\gets} \nabla_{\theta} \mathcal{L}_{\text{VAE}}(\theta)$
\STATE \hspace{0.6cm}$ \phi \bar{\gets} \nabla_{\phi} \mathcal{L}_{\text{D}}(\phi)$
\STATE \hspace{0.3cm}$ \textbf{end while} $
\end{algorithmic}
\end{algorithm}

DisCoPatch~(our proposed method) is an Adversarial VAE, which is composed of a VAE and a Discriminator. The VAE features an Encoder ($\mathcal{E}_{\theta_{\text{E}}}$), consisting of convolutional layers with a kernel size of 3, stride 2, padding 1, and output padding of 1. All the convolution layers are followed by BN and a \texttt{LeakyReLU} activation function. The number of filters doubles with each layer. Encoded features are then flattened and passed through two distinct fully connected layers, one estimating $z_{\mu}$ and the other $z_{\sigma}$, with outputs the size of the latent dimension. These outputs undergo the reparametrization trick to generate $z$, which is then fed into the VAE's decoder, referred to as the Generator ($\mathcal{G}_{\theta_{\text{G}}}$). The Generator comprises transposed convolutions, followed by BN and a \texttt{LeakyReLU} activation, with the same kernel size, stride, padding, and output padding as the Encoder. The number of filters halves after each layer. A final convolutional layer with a kernel size of 3 and padding of 1, followed by a \texttt{Tanh} activation, generates the final output image. The generated image is subsequently fed into a Discriminator ($\mathcal{D}_{\phi}$). The Discriminator shares the same architecture as the Encoder but replaces the two fully connected layers with a single one that generates an output of size 1, followed by a \texttt{Sigmoid} activation. Additionally, for its recommended setup, \texttt{track\_running\_stats} is set to \texttt{False} in the Discriminator. The training process is covered in detail in Subsection~\ref{subsec:training}, but can be summarized by Algorithm~\ref{alg:train}.

\begin{table}[!ht]
    \centering
    \caption{Hyperparameters used for DisCoPatch's training.}
    
    \label{tab:DisCoPatch_imp}
    {\resizebox{1.0\linewidth}{!}{
    \begin{tabular}{lccc}
        \toprule
        \textbf{Model} & \textbf{Lat. Dim.} & \textbf{Hidden Dimensions} & \textbf{lr} \\
        \toprule 
        Full-Size      & 1024     & 32, 64, 128, 256, 512, 1024 & $5e^{-5}$ \\
        Patches        & 1024     & 128, 256, 512, 1024         & $8.5e^{-5}$\\
        \bottomrule
    \end{tabular}}}
\end{table}

DisCoPatch is optimized using the Adam optimizer, with $\beta_1=0.9$ and $\beta_2=0.999$. Both models share the same learning rate, $lr$. As shown in Equation~\ref{eq:vae_loss_final}, three weighing terms are required to train the model; these were fixed for all datasets, with $\omega_{\text{KL}} = 1e^{-4}$, $\omega_{\text{Rec}} = 1e^{-3}$ and $\omega_{\text{Gen}} = 1e^{-3}$. Additional hyperparameters can be found in Table~\ref{tab:DisCoPatch_imp}. The developed code is based on a publicly available repository~\footnote{\href{https://github.com/AntixK/PyTorch-VAE}{https://github.com/AntixK/PyTorch-VAE}}. The referred repository is released under the Apache 2.0 License.

\section{Compute Resources}
\label{subsec:compute_sup}

This appendix describes the computational resources employed for inference in the selected models and to train DisCoPatch.

\subsection{Training}

DisCoPatch was trained on ImageNet-1K using a system equipped with an NVIDIA H100 Tensor Core GPU~(94 GB VRAM), a 32-core, 64-thread AMD EPYC 9334 CPU, and 768 GB RAM. More information can be found in Table~\ref{tab:comp}.

\begin{table}[!ht]
    \centering
    \caption{Summary of the compute resources required for training the DisCoPatch models on ImageNet-1K.}
    
    \label{tab:comp}
    {\resizebox{1.0\linewidth}{!}{
    \begin{tabular}{lccccc}
        \toprule
        \textbf{Model} & \textbf{\#Patches} & \textbf{Batch Size} & \textbf{Epochs} & \textbf{Parameters} & \textbf{Time~(s)}\\
        \midrule
        Full-Size & 1       & 660                 & 70             & 69,240,517   & 111,471\\
        Patches   & 48        & 67                  & 30             & 69,118,340         & 94,323 \\
        \bottomrule
    \end{tabular}}}
\end{table}

\subsection{Inference}

To measure the models' latency, we fed them 1000 individual inputs and calculated the average inference time per image. Table~\ref{tab:inference} reveals DisCoPatch has the lowest latency. Although MobileNetV2 is smaller, it takes 224$\times$224 inputs, whereas DisCoPatch processes an image as a batch of 64$\times$64 patches. Latency was measured on a machine with an NVIDIA RTX4070 GPU (8 GB VRAM), an 8-core, 16-thread AMD RYZEN 9 8945HS CPU, and 32 GB RAM.

\begin{table}[!ht]
    \centering
    \caption{Latency of the tested models.}
    \label{tab:inference}
    {\resizebox{0.9\linewidth}{!}{
    \begin{tabular}{lcc}
        \toprule
        \textbf{Model} & \textbf{\#Parameters} & \textbf{Latency (ms)} \\
        \toprule
                               MOODv2~(BEiTv2)                 & 86,530,984  & 19.26\\
                               SCALE~(ResNet-50)                &  25,557,032   & 11.27 \\
                               NNGuide~(RegNet)       &  83,590,140  & 31.00 \\
                               NNGuide~(ResNet-50)       &   25,557,032   & 11.10 \\
                               NNGuide~(MobileNetv2)  &   \textbf{3,504,872}   & 9.40 \\
                               RankFeat~(ResNetv2-101) & 44,549,160 & 15.05\\
                               ASH~(ResNet-50) & 25,557,032  & 11.15 \\
                               FDBD~(ResNet-50) & 25,557,032  & 11.17 \\
                               DisCoPatch-64           &  6,218,753     & \textbf{1.56} \\
        \bottomrule
    \end{tabular}}}
\end{table}

\section{Batch Normalization Bias}
\label{subsec:batch_norm_bias}

Table~\ref{tab:batch_bias} reveals a critical limitation in the model’s behavior when evaluated with the BatchNorm employed in its standard \emph{evaluation mode}: the model fails to distinguish between ID and OOD samples reliably. In contrast, when disabling the use of the learned statistics and instead using batch-specific statistics, the model's performance improves significantly, even with a batch size of 1. This effect demonstrates that the running statistics acquired during training are ineffective for discriminating ID from OOD, while the \emph{test} batch statistics provide more discriminating power for detecting OOD samples. It should be noted that as the batch size increases, this improvement becomes more pronounced, which indicates that the model has developed a dependency/shortcut on batch-specific statistics, instead of leveraging the running mean and variance acquired during training. This means that the use of BatchNorm's running statistics compromises robustness, as it has been observed in adversarial and OOD scenarios~\cite{benz2021batch, wang2022removing}.

\begin{table}[!ht]
    \centering
    \caption{OOD detection performance, reported as AUROC/FPR95, of a DisCoPatch model trained on complete ImageNet-1k images. Legend: BS~=~Batch Size.}
    \label{tab:batch_bias}
    {\resizebox{0.8\linewidth}{!}{
    \begin{tabular}{lccc}
    \toprule
     \multicolumn{2}{l}{\textbf{Mode}} & \textbf{Near-OOD} & \textbf{Far-OOD} \\
     \midrule
      \multicolumn{2}{l}{Learned Statistics} & 38.4/98.0 & 34.5/98.1\\
      \midrule
      \multirow{6}{*}{Batch Statistics} & BS=1 & 64.9/87.4 & 70.9/71.2 \\
      & BS=16 & 90.2/55.5 & 91.1/40.0 \\
      & BS=32 & 95.7/28.6 & 93.4/36.2 \\
      & BS=64 & 99.3/2.2 & 96.1/23.8 \\
      & BS=128 & 99.8/0.3 & 97.4/17.0 \\
      & BS=256 & \textbf{100.0/0.0} & \textbf{98.1/12.5} \\
      \bottomrule
    \end{tabular}}}
\end{table}

It is important to note that in the configuration employed for this experiment, each batch contains exclusively ID or OOD samples. This means that a single anomaly score predicted for an image by the \emph{Batch Statistics} mode is dependent on the statistics from every image in the batch. This design constraint limits the suitability of this configuration for multiple applications because it requires that all images in a batch share the same class type. A practical and effective solution to ensure this homogeneity without prior class knowledge is by constructing each batch from patches of the same image.

\section{Alternative Normalization Layers}
\label{subsec:alt_sup}

To better demonstrate the effect of the BatchNorm layer and batch statistics, we replaced BatchNorm with GroupNorm and InstanceNorm and retrained DisCoPatch. As shown in Table~\ref{tab:norm}, GroupNorm performs the worst, while InstanceNorm shows intermediate results. These findings suggest that avoiding normalization across multiple channels is beneficial. However, BatchNorm consistently achieves the best performance, particularly under Covariate Shift, highlighting the importance of batch-level statistics.

\begin{table}[!ht]
    \centering
    \caption{Comparison between different normalization layers on DisCoPatch. The results correspond to the 64-patch configuration.}
    \label{tab:norm}
    {\resizebox{\linewidth}{!}{
    \begin{tabular}{@{\extracolsep{2pt}}lccc@{}}
        \toprule
         \textbf{Normalization} & \multicolumn{1}{c}{\textbf{Near-OOD}} & \multicolumn{1}{c}{\textbf{Far-OOD}} & \textbf{Covariate Shift OOD}\\
         \midrule
         GroupNorm & 85.0 & 81.0 & 78.5 \\
         InstanceNorm & \underline{92.8} & \underline{93.7} & \underline{86.2} \\
         BatchNorm & \phantom{\textsuperscript{+0.0}}\textbf{95.1}\textsuperscript{\textcolor{darkgreen}{+2.3}} & \phantom{\textsuperscript{+0.0}}\textbf{96.6}\textsuperscript{\textcolor{darkgreen}{+2.9}} & \phantom{\textsuperscript{+00.0}}\textbf{97.2}\textsuperscript{\textcolor{darkgreen}{+11.0}} \\
         \bottomrule
    \end{tabular}}}
\end{table}

\section{Patch Count Imapct}
\label{subsec:count_sup}

Table~\ref{tab:openood1k-table-patch} shows that the use of more patches improves the detection accuracy, reaching a plateau around 64 patches.

\begin{table}[!ht]
    \centering
    \caption{OOD detection results on ImageNet-1K for multiple patch counts.}
    \label{tab:openood1k-table-patch}
    {\resizebox{\linewidth}{!}{
    \begin{tabular}{@{\extracolsep{2pt}}lccc@{}}
        \toprule
         \textbf{Model} & \multicolumn{1}{c}{\textbf{Near-OOD}} & \multicolumn{1}{c}{\textbf{Far-OOD}} & \textbf{Covar. Shift OOD}\\
         \midrule
         MOODv2~(BEiTv2) & 88.9 & \underline{97.1} & 70.5 \\
         SCALE~(ResNet-50) & 81.4 & 96.5 & 83.3 \\
         NNGuide~(RegNet) & 89.2 & \textbf{97.8} & 78.5\\
         RankFeat~\cite{song2024rankfeat_short}~(ResNetv2-101) & \underline{89.7} & 94.5 & \underline{91.9} \\
         ASH~\cite{djurisic2022extremely_short}~(ResNet-50)& 78.6 & 96.1 & 84.7 \\
         FDBD~\cite{liu2023fast_short}~(ResNet-50)& 74.9 & 95.8 & 82.2 \\
         DisCoPatch-4~(Proposed) & \phantom{\textsuperscript{+0.0}}90.3\textsuperscript{\textcolor{darkgreen}{+0.6}} & \phantom{\textsuperscript{-0.0}}92.6\textsuperscript{\textcolor{red}{-5.2}} & \phantom{\textsuperscript{+0.0}}93.0\textsuperscript{\textcolor{darkgreen}{+1.1}} \\
         DisCoPatch-16~(Proposed) & \phantom{\textsuperscript{+0.0}}94.3\textsuperscript{\textcolor{darkgreen}{+4.6}} & \phantom{\textsuperscript{-0.0}}96.0\textsuperscript{\textcolor{red}{-1.8}} & \phantom{\textsuperscript{+0.0}}96.5\textsuperscript{\textcolor{darkgreen}{+3.6}} \\
         DisCoPatch-32~(Proposed) & \phantom{\textsuperscript{+0.0}}94.8\textsuperscript{\textcolor{darkgreen}{+5.1}} & \phantom{\textsuperscript{-0.0}}96.4\textsuperscript{\textcolor{red}{-1.4}} & \phantom{\textsuperscript{+0.0}}96.9\textsuperscript{\textcolor{darkgreen}{+4.0}} \\
         DisCoPatch-64~(Proposed) & \phantom{\textsuperscript{+0.0}}95.1\textsuperscript{\textcolor{darkgreen}{+5.4}} & \phantom{\textsuperscript{-0.0}}96.6\textsuperscript{\textcolor{red}{-1.2}} & \phantom{\textsuperscript{+0.0}}97.2\textsuperscript{\textcolor{darkgreen}{+4.3}} \\
         DisCoPatch-128~(Proposed) & \phantom{\textsuperscript{+0.0}}95.2\textsuperscript{\textcolor{darkgreen}{+5.5}} & \phantom{\textsuperscript{-0.0}}96.7\textsuperscript{\textcolor{red}{-1.1}} & \phantom{\textsuperscript{+0.0}}97.3\textsuperscript{\textcolor{darkgreen}{+4.4}} \\
         DisCoPatch-512~(Proposed) & \phantom{\textsuperscript{+0.0}}\textbf{95.3}\textsuperscript{\textcolor{darkgreen}{+5.6}} & \phantom{\textsuperscript{-0.0}}96.8\textsuperscript{\textcolor{red}{-1.0}} & \phantom{\textsuperscript{+0.0}}\textbf{97.4}\textsuperscript{\textcolor{darkgreen}{+4.5}} \\
         \bottomrule
    \end{tabular}}}
\end{table}

\section{Covariate Shift Results on ImageNet-1K}
\label{subsec:results_sup_in1k}

This appendix contains the performance metrics per corruption achieved on the Covariate Shift OOD benchmark.

\subsection{MOODv2~(BEiTv2)}

MOODv2 obtains its best results for corruptions that filter high-frequency components. Table~\ref{tab:in1k-covar-mood} shows that it scores very low for Intensity 1 in all corruption tests.

\begin{table}[!ht]
    \centering
    \caption{Covariate shift OOD benchmark for MOODv2.}
    \label{tab:in1k-covar-mood}
    {\resizebox{1.0\linewidth}{!}{
    \begin{tabular}{@{\extracolsep{0pt}}lcccccc@{}}
        \toprule
        \multirow{2}{*}{\textbf{Corruption}} & \multicolumn{5}{c}{\textbf{Corruption Intensity}} & \multirow{2}{*}{\textbf{Average}}\\
        \cline{2-6}
        & \textbf{1} & \textbf{2} & \textbf{3} & \textbf{4} & \textbf{5} & \\
         \toprule
        Brightness & 54.2/93.8 & 55.2/93.2 & 56.9/92.2 & 59.2/90.6 & 62.2/88.2 & 57.5/91.6 \\
        Contrast & 59.6/91.6 & 61.7/90.4 & 66.1/87.7 & 77.5/76.6 & 86.4/54.9 & 70.3/80.2 \\
        Defocus Blur & 69.6/80.3 & 76.0/70.7 & 85.5/51.0 & 91.8/33.9 & 95.4/21.0 & 83.7/51.4 \\
        Elastic Transform & 60.2/88.5 & 75.9/62.7 & 63.2/85.1 & 70.8/75.3 & 87.1/44.2 & 71.4/71.1 \\
        Fog & 70.0/82.5 & 77.3/71.5 & 89.2/40.0 & 93.3/25.6 & 97.0/11.4 & 85.3/46.2 \\
        Frost & 61.8/88.8 & 70.3/79.9 & 75.6/71.7 & 77.1/69.7 & 80.0/63.5 & 72.9/74.7 \\
        Glass Blur & 60.6/90.6 & 72.1/78.0 & 81.6/62.0 & 88.9/44.4 & 96.5/17.6 & 79.9/58.5 \\
        Gaussian Blur & 58.0/90.0 & 60.0/87.7 & 64.8/82.8 & 71.5/74.1 & 80.3/59.0 & 66.9/78.7 \\
        Gaussian Noise & 63.2/86.3 & 70.2/77.8 & 83.9/51.0 & 87.9/41.4 & 93.3/26.5 & 79.7/56.6 \\
        Impulse Noise & 57.2/89.8 & 60.4/86.8 & 63.6/83.5 & 71.0/74.6 & 78.9/61.6 & 66.2/79.3 \\
        JPEG Compression & 63.5/88.0 & 65.9/85.7 & 67.6/83.8 & 71.9/77.6 & 77.5/68.5 & 69.2/80.7 \\
        Motion Blur & 58.7/90.2 & 63.1/85.9 & 70.5/77.0 & 80.0/61.0 & 85.8/48.5 & 71.6/72.5 \\
        Pixelate & 55.8/92.2 & 57.5/90.9 & 61.0/87.9 & 67.7/81.0 & 83.8/57.1 & 65.2/81.8 \\
        Saturate & 54.0/93.4 & 55.6/92.2 & 55.4/93.1 & 60.3/90.0 & 65.2/85.2 & 58.1/90.8 \\
        Shot Noise & 58.3/89.7 & 61.0/86.8 & 65.3/82.1 & 73.7/70.7 & 80.1/59.1 & 67.7/77.7 \\
        Snow & 62.3/87.4 & 70.8/77.8 & 70.7/78.9 & 75.6/71.4 & 77.2/67.7 & 71.3/76.6 \\
        Spatter & 55.3/92.9 & 59.1/90.2 & 62.0/87.8 & 64.2/85.7 & 69.4/80.0 & 62.0/87.3 \\
        Speckle Noise & 57.6/90.4 & 59.1/88.8 & 63.9/83.7 & 67.4/79.3 & 72.3/72.0 & 64.1/82.8 \\
        Zoom Blur & 65.8/83.7 & 71.9/75.5 & 76.7/67.1 & 81.3/58.3 & 86.5/46.2 & 76.5/66.1 \\
        \bottomrule
        \addlinespace[0.5mm]
        \textbf{Average}  & 60.3/88.9 & 65.4/82.8 & 69.7/76.2 & 75.3/67.4 & 81.8/54.3 & 70.5/73.9 \\  
         \bottomrule
    \end{tabular}}}
\end{table}

\subsection{SCALE~(ResNet-50)}

The results in Table~\ref{tab:in1k-covar-scale} demonstrate a drop in performance for Intensity 2 in some corruptions that dampen the high-frequency components, such as blurs. This occurs despite good performance on intensity 1.

\begin{table}[!ht]
    \centering
    \caption{Covariate shift OOD benchmark for SCALE.}
    \label{tab:in1k-covar-scale}
    {\resizebox{1.0\linewidth}{!}{
    \begin{tabular}{@{\extracolsep{0pt}}lcccccc@{}}
        \toprule
        \multirow{2}{*}{\textbf{Corruption}} & \multicolumn{5}{c}{\textbf{Corruption Intensity}} & \multirow{2}{*}{\textbf{Average}}\\
        \cline{2-6}
        & \textbf{1} & \textbf{2} & \textbf{3} & \textbf{4} & \textbf{5} & \\
         \toprule
        Brightness & 60.1/90.3 & 61.2/89.7 & 63.9/88.0 & 68.3/84.8 & 74.1/79.0 & 65.5/86.4 \\
        Contrast & 74.5/76.9 & 80.3/67.7 & 88.6/48.6 & 98.1/9.6 & 99.9/0.3 & 88.3/40.6 \\
        Defocus Blur & 82.3/63.6 & 87.3/52.4 & 93.9/31.5 & 97.1/15.5 & 98.8/6.1 & 91.9/33.8 \\
        Elastic Transform & 71.1/80.3 & 82.5/62.5 & 81.3/62.9 & 87.9/46.7 & 95.0/19.8 & 83.6/54.4 \\
        Fog & 83.4/53.9 & 73.1/79.0 & 77.6/72.7 & 83.2/62.3 & 86.9/53.1 & 80.8/64.2 \\
        Frost & 93.7/29.9 & 72.5/80.0 & 85.2/59.6 & 91.3/41.1 & 92.2/38.0 & 87.0/49.7 \\
        Gaussian Blur & 94.7/26.8 & 74.6/76.3 & 85.0/57.8 & 92.3/36.9 & 96.1/20.9 & 88.6/43.7 \\
        Gaussian Noise & 99.1/4.3 & 66.8/86.4 & 76.0/77.3 & 87.7/54.1 & 95.8/21.9 & 85.1/48.8 \\
        Glass Blur & 99.3/3.5 & 82.4/62.0 & 89.9/43.1 & 97.2/14.7 & 98.2/9.4 & 93.4/26.5 \\
        Impulse Noise & 99.0/4.7 & 75.6/77.7 & 83.2/65.6 & 89.0/50.0 & 96.7/16.9 & 88.7/43.0 \\
        JPEG Compression & 99.2/3.5 & 70.7/81.0 & 73.0/78.6 & 74.8/76.6 & 81.3/67.8 & 79.8/61.5 \\
        Motion Blur & 89.1/48.9 & 73.7/78.0 & 81.6/66.5 & 90.9/43.5 & 96.6/18.2 & 86.4/51.0 \\
        Pixelate & 98.3/8.3 & 65.0/86.9 & 64.9/87.1 & 79.5/70.6 & 89.6/46.3 & 79.4/59.8 \\
        Saturate & 93.4/32.2 & 65.3/86.9 & 64.1/88.6 & 61.4/89.6 & 68.4/85.2 & 70.5/76.5 \\
        Shot Noise & 76.0/77.3 & 68.0/85.2 & 77.8/74.1 & 88.0/52.3 & 96.5/17.8 & 81.3/61.3 \\
        Snow & 98.7/6.5 & 73.0/80.2 & 88.5/50.2 & 85.8/59.7 & 91.5/40.8 & 87.5/47.5 \\
        Spatter & 95.2/24.7 & 61.5/89.4 & 69.2/83.6 & 76.3/75.5 & 79.0/70.4 & 76.2/68.7 \\
        Speckle Noise & 84.1/58.1 & 66.8/85.9 & 70.9/82.3 & 83.7/63.0 & 89.5/47.3 & 79.0/67.3 \\
        Zoom Blur & 94.2/29.0 & 81.9/64.0 & 87.3/51.2 & 90.4/42.3 & 93.0/32.9 & 89.3/43.9 \\
        \bottomrule
        \addlinespace[0.5mm]
        \textbf{Average}  & 88.7/38.0 & 72.7/77.4 & 79.0/66.8 & 85.4/52.0 & 90.5/36.4 & 83.3/54.1 \\  
         \bottomrule
    \end{tabular}}}
\end{table}

\subsection{NNGuide~(RegNet)}

NNGuide surpasses the performance of MOODv2, as demonstrated by the results in Table~\ref{tab:in1k-covar-nnguide_1}, particularly for higher corruption intensities. Nonetheless, it also suffers from significantly low scores at Intensity 1.

\begin{table}[!ht]
    \centering
    \caption{Covariate shift OOD benchmark for NNGuide~(RegNet).}
    \label{tab:in1k-covar-nnguide_1}
    {\resizebox{1.0\linewidth}{!}{
    \begin{tabular}{@{\extracolsep{0pt}}lcccccc@{}}
        \toprule
        \multirow{2}{*}{\textbf{Corruption}} & \multicolumn{5}{c}{\textbf{Corruption Intensity}} & \multirow{2}{*}{\textbf{Average}}\\
        \cline{2-6}
        & \textbf{1} & \textbf{2} & \textbf{3} & \textbf{4} & \textbf{5} & \\
         \toprule
        Brightness & 57.9/91.7 & 59.8/90.9 & 62.8/89.2 & 67.3/85.7 & 73.0/79.8 & 64.2/87.5 \\
        Contrast & 75.8/76.1 & 82.3/64.5 & 91.4/37.9 & 98.7/6.1 & 99.9/0.4 & 89.6/37.0 \\
        Defocus Blur & 72.1/75.6 & 78.1/66.0 & 87.9/43.9 & 93.7/26.2 & 96.9/14.2 & 85.7/45.2 \\
        Elastic Transform & 66.2/84.8 & 78.0/66.5 & 73.4/75.0 & 82.7/57.8 & 94.6/23.5 & 79.0/61.5 \\
        Fog & 75.9/75.9 & 81.2/66.3 & 88.2/48.5 & 91.7/37.3 & 96.1/19.6 & 86.6/49.5 \\
        Frost & 70.8/81.4 & 81.7/64.0 & 87.8/49.1 & 88.7/46.7 & 91.6/37.0 & 84.1/55.6 \\
        Gaussian Blur & 64.0/86.4 & 75.0/71.3 & 83.9/54.6 & 90.0/38.8 & 96.7/15.2 & 81.9/53.3 \\
        Gaussian Noise & 64.0/86.6 & 70.3/79.9 & 80.5/63.8 & 90.8/37.6 & 98.0/9.5 & 80.7/55.5 \\
        Glass Blur & 71.2/77.9 & 80.3/62.8 & 93.4/27.2 & 95.7/18.6 & 97.4/11.5 & 87.6/39.6 \\
        Impulse Noise & 62.2/87.3 & 68.2/80.9 & 74.2/72.8 & 87.4/46.2 & 96.6/15.7 & 77.7/60.6 \\
        JPEG Compression & 60.9/89.0 & 63.7/86.7 & 65.9/84.9 & 71.7/78.3 & 78.8/68.0 & 68.2/81.4 \\
        Motion Blur & 62.6/86.7 & 69.0/79.6 & 79.3/63.5 & 89.4/39.7 & 93.9/25.3 & 78.8/59.0 \\
        Pixelate & 62.3/88.5 & 63.8/87.1 & 68.1/82.4 & 75.4/72.1 & 82.1/60.2 & 70.3/78.0 \\
        Saturate & 61.3/89.7 & 62.9/88.2 & 58.5/91.6 & 68.7/84.7 & 77.7/72.1 & 65.8/85.2 \\
        Shot Noise & 66.0/85.1 & 72.9/76.8 & 81.1/62.0 & 92.7/30.6 & 97.3/12.3 & 82.0/53.4 \\
        Snow & 69.6/82.5 & 80.9/64.4 & 80.1/65.5 & 86.4/51.1 & 91.0/37.5 & 81.6/60.2 \\
        Spatter & 62.1/89.4 & 67.7/85.7 & 70.5/82.3 & 74.7/78.2 & 81.6/66.4 & 71.3/80.4 \\
        Speckle Noise & 63.8/87.4 & 66.8/84.4 & 76.6/70.4 & 82.5/58.4 & 89.2/41.3 & 75.8/68.4 \\
        Zoom Blur & 70.1/79.0 & 76.7/69.1 & 82.7/57.8 & 86.5/48.6 & 90.4/38.0 & 81.3/58.5 \\
        \bottomrule
        \addlinespace[0.5mm]
        \textbf{Average} & 66.3/84.3 & 72.6/75.5 & 78.2/64.3 & 85.0/49.6 & 90.7/34.1 & 78.5/61.6 \\
         \bottomrule
    \end{tabular}}}
\end{table}

\newpage

\subsection{NNGuide~(ResNet-50)}

The behavior observed for this backbone of NNGuide in Table~\ref{tab:in1k-covar-nnguide_2} is very similar to the one observed when the RegNet was used. However, it is slightly more effective than the bigger backbone at Covariate Shift detection.

\begin{table}[!ht]
    \centering
    \caption{Covariate shift OOD benchmark for NNGuide~(ResNet-50).}
    \label{tab:in1k-covar-nnguide_2}
    {\resizebox{1.0\linewidth}{!}{
    \begin{tabular}{@{\extracolsep{0pt}}lcccccc@{}}
        \toprule
        \multirow{2}{*}{\textbf{Corruption}} & \multicolumn{5}{c}{\textbf{Corruption Intensity}} & \multirow{2}{*}{\textbf{Average}}\\
        \cline{2-6}
        & \textbf{1} & \textbf{2} & \textbf{3} & \textbf{4} & \textbf{5} & \\
         \toprule
        Brightness & 56.5/90.0 & 58.8/88.5 & 62.7/85.6 & 68.3/80.5 & 74.7/72.6 & 64.2/83.4 \\
        Contrast & 54.1/91.3 & 61.9/86.2 & 75.0/72.1 & 93.3/27.8 & 99.2/3.0 & 76.7/56.1 \\
        Defocus Blur & 74.7/69.2 & 81.2/57.6 & 90.3/36.5 & 94.7/22.6 & 97.2/13.0 & 87.6/39.8 \\
        Elastic Transform & 64.8/83.3 & 80.4/65.4 & 77.6/66.9 & 86.4/51.0 & 95.7/22.1 & 81.0/57.7 \\
        Fog & 69.5/79.0 & 74.6/73.2 & 80.7/63.9 & 84.5/55.3 & 92.2/32.8 & 80.3/60.8 \\
        Frost & 71.7/73.9 & 84.8/49.8 & 91.0/33.4 & 91.7/31.0 & 94.2/22.7 & 86.7/42.2 \\
        Gaussian Blur & 50.3/93.5 & 65.9/81.8 & 77.6/67.9 & 85.9/51.7 & 94.7/23.6 & 74.9/63.7 \\
        Gaussian Noise & 66.2/82.5 & 74.4/73.1 & 84.9/53.6 & 93.8/27.1 & 98.6/6.4 & 83.6/48.5 \\
        Glass Blur & 76.6/68.1 & 86.4/47.7 & 96.1/16.5 & 97.6/10.5 & 98.6/6.2 & 91.1/29.8 \\
        Impulse Noise & 78.1/67.9 & 82.3/61.0 & 85.9/52.5 & 93.8/27.6 & 98.3/8.2 & 87.7/43.5 \\
        JPEG Compression & 63.4/84.7 & 66.6/81.5 & 69.3/78.4 & 77.6/66.5 & 87.2/46.5 & 72.8/71.5 \\
        Motion Blur & 69.3/76.6 & 78.8/62.3 & 88.7/41.2 & 94.9/22.1 & 97.0/14.0 & 85.7/43.2 \\
        Pixelate & 63.3/86.5 & 65.2/84.9 & 75.6/71.9 & 86.9/48.2 & 92.2/32.1 & 76.7/64.7 \\
        Saturate & 47.3/96.3 & 49.9/95.2 & 45.5/96.9 & 57.2/94.6 & 66.5/91.2 & 53.3/94.8 \\
        Shot Noise & 68.1/80.7 & 77.2/69.5 & 86.1/51.7 & 95.0/23.1 & 97.9/10.3 & 84.9/47.1 \\
        Snow & 74.9/74.5 & 89.2/42.3 & 87.6/48.4 & 92.8/31.5 & 95.2/21.5 & 87.9/43.7 \\
        Spatter & 43.5/97.1 & 55.6/95.7 & 65.0/93.6 & 71.2/91.4 & 78.2/87.1 & 62.7/93.0 \\
        Speckle Noise & 52.0/95.1 & 57.2/93.2 & 71.5/84.2 & 78.6/75.7 & 85.4/62.5 & 68.9/82.2 \\
        Zoom Blur & 77.2/68.8 & 83.4/57.8 & 87.6/47.8 & 90.5/39.5 & 92.9/31.3 & 86.3/49.0 \\
        \bottomrule
        \addlinespace[0.5mm]
        \textbf{Average} & 64.3/82.0 & 72.3/71.9 & 78.9/61.2 & 86.0/46.2 & 91.4/32.0 & 78.6/58.7 \\ 
         \bottomrule
    \end{tabular}}}
\end{table}

\subsection{NNGuide~(MobileNet)}

The behavior observed for this backbone of NNGuide in Table~\ref{tab:in1k-covar-nnguide_3} is very similar to the one observed when the RegNet and ResNet were used. Its performance is lower than the one achieved by the other backbones but by a small margin.

\begin{table}[!ht]
    \centering
    \caption{Covariate shift OOD benchmark for NNGuide~(MobileNet).}
    \label{tab:in1k-covar-nnguide_3}
    {\resizebox{1.0\linewidth}{!}{
    \begin{tabular}{@{\extracolsep{0pt}}lcccccc@{}}
        \toprule
        \multirow{2}{*}{\textbf{Corruption}} & \multicolumn{5}{c}{\textbf{Corruption Intensity}} & \multirow{2}{*}{\textbf{Average}}\\
        \cline{2-6}
        & \textbf{1} & \textbf{2} & \textbf{3} & \textbf{4} & \textbf{5} & \\
         \toprule
        Brightness & 52.6/94.1 & 54.0/93.6 & 56.6/92.7 & 60.8/90.7 & 66.0/87.2 & 58.0/91.7 \\
        Contrast & 55.9/93.0 & 58.3/91.7 & 62.9/88.2 & 76.5/70.7 & 91.6/32.5 & 69.1/75.2 \\
        Defocus Blur & 68.2/82.8 & 76.3/71.5 & 88.6/44.7 & 94.4/24.5 & 96.8/14.7 & 84.9/47.6 \\
        Elastic Transform & 58.2/92.1 & 73.6/79.3 & 70.5/84.0 & 80.8/71.2 & 90.9/48.9 & 74.8/75.1 \\
        Fog & 59.1/92.1 & 62.6/90.2 & 69.5/84.0 & 75.9/74.6 & 87.1/51.2 & 70.8/78.4 \\
        Frost & 63.8/89.7 & 76.5/77.6 & 83.2/66.6 & 84.4/64.0 & 87.4/56.6 & 79.0/70.9 \\
        Gaussian Blur & 58.0/91.4 & 72.9/76.1 & 86.3/50.0 & 93.9/26.2 & 98.1/9.2 & 81.8/50.6 \\
        Gaussian Noise & 68.0/82.9 & 80.1/63.4 & 93.3/27.2 & 98.6/6.1 & 99.8/0.9 & 88.0/36.1 \\
        Glass Blur & 71.8/80.4 & 83.1/59.6 & 94.9/22.6 & 97.0/14.0 & 98.4/7.3 & 89.0/36.8 \\
        Impulse Noise & 72.9/75.6 & 83.5/57.3 & 90.9/36.1 & 98.5/6.6 & 99.8/0.8 & 89.1/35.3 \\
        JPEG Compression & 60.0/92.3 & 62.3/91.6 & 64.0/90.7 & 69.3/87.1 & 76.2/78.8 & 66.4/88.1 \\
        Motion Blur & 62.4/88.6 & 73.6/75.8 & 86.6/48.3 & 94.7/21.9 & 97.0/13.0 & 82.9/49.5 \\
        Pixelate & 61.5/91.1 & 67.2/86.7 & 76.1/73.3 & 90.2/37.2 & 94.2/22.3 & 77.9/62.1 \\
        Saturate & 55.0/94.0 & 56.0/93.4 & 55.8/92.4 & 65.3/86.6 & 73.0/78.4 & 61.0/89.0 \\
        Shot Noise & 68.7/82.5 & 82.1/59.6 & 93.1/28.3 & 98.7/6.0 & 99.6/1.8 & 88.4/35.6 \\
        Snow & 71.7/80.7 & 85.4/59.1 & 82.6/65.2 & 88.1/52.1 & 90.5/45.6 & 83.7/60.5 \\
        Spatter & 55.4/92.7 & 69.2/82.3 & 76.1/77.2 & 81.1/66.8 & 85.8/60.1 & 73.5/75.8 \\
        Speckle Noise & 64.8/87.2 & 70.7/80.6 & 86.1/50.9 & 91.9/33.2 & 96.0/18.0 & 81.9/54.0 \\
        Zoom Blur & 74.0/76.5 & 81.5/63.6 & 86.1/53.6 & 89.4/43.9 & 91.8/36.0 & 84.6/54.7 \\
        \bottomrule
        \addlinespace[0.5mm]
        \textbf{Average} & 63.3/87.3 & 72.0/76.5 & 79.1/61.9 & 85.8/46.5 & 90.5/34.9 & 78.1/61.4 \\
         \bottomrule
    \end{tabular}}}
\end{table}

\subsection{RankFeat~(ResNetv2-101)}

The performance achieved by RankFeat in Table~\ref{tab:in1k-covar-rankfeat} is significantly superior to that achieved by the other baseline methods. Nevertheless, it remains inferior to DisCoPatch in terms of Covariate Shift OOD detection abilities.

\begin{table}[!ht]
    \centering
    \caption{Covariate shift OOD benchmark for RankFeat.}
    \label{tab:in1k-covar-rankfeat}
    {\resizebox{1.0\linewidth}{!}{
    \begin{tabular}{@{\extracolsep{0pt}}lcccccc@{}}
        \toprule
        \multirow{2}{*}{\textbf{Corruption}} & \multicolumn{5}{c}{\textbf{Corruption Intensity}} & \multirow{2}{*}{\textbf{Average}}\\
        \cline{2-6}
        & \textbf{1} & \textbf{2} & \textbf{3} & \textbf{4} & \textbf{5} & \\
         \toprule
        Brightness & 80.9/67.3 & 82.6/64.2 & 84.7/60.0 & 87.1/54.5 & 89.6/47.9 & 85.0/58.8 \\
        Contrast & 83.2/63.4 & 84.7/60.6 & 87.3/54.9 & 92.8/37.6 & 97.4/14.3 & 89.1/46.1 \\
        Defocus Blur & 88.1/52.9 & 90.4/46.5 & 93.0/39.4 & 94.7/31.6 & 96.1/23.7 & 92.5/38.8 \\
        Elastic Transform & 84.0/61.8 & 87.6/56.2 & 88.7/44.3 & 90.8/37.9 & 95.4/22.1 & 89.3/44.4 \\
        Fog & 85.1/60.3 & 86.8/56.8 & 89.3/50.9 & 90.8/46.0 & 93.6/33.9 & 89.1/49.6 \\
        Frost & 89.1/43.6 & 94.2/25.2 & 96.4/16.2 & 96.7/14.9 & 97.6/10.8 & 94.8/22.1 \\
        Gaussian Blur & 83.7/62.6 & 89.1/51.0 & 91.9/43.8 & 93.7/38.0 & 96.3/22.3 & 91.0/43.5\\
        Gaussian Noise & 87.8/52.3 & 91.1/42.2 & 94.4/28.8 & 96.9/15.8 & 98.7/5.4 & 93.8/28.9 \\
        Glass Blur & 90.9/38.7 & 93.9/27.9 & 97.1/14.5 & 97.4/13.4 & 97.0/16.9 & 95.3/22.3 \\
        Impulse Noise & 94.0/31.3 & 95.4/24.9 & 96.2/20.2 & 97.7/11.2 & 98.8/4.7 & 96.4/18.5 \\
        JPEG Compression & 84.7/61.7 & 86.6/57.9 & 87.9/55.0 & 91.1/45.8 & 93.5/37.1 & 88.7/51.5 \\
        Motion Blur & 84.9/59.0 & 88.5/50.5 & 92.7/37.9 & 96.0/22.7 & 97.4/14.6 & 91.9/36.9 \\
        Pixelate & 84.6/62.3 & 85.2/62.8 & 90.4/48.3 & 95.4/26.3 & 97.5/12.9 & 90.6/42.5 \\
        Saturate & 81.4/67.6 & 81.2/68.8 & 82.6/64.9 & 88.4/52.1 & 91.1/43.4 & 84.9/59.3\\
        Shot Noise & 88.2/50.4 & 91.7/39.1 & 94.5/27.4 & 97.3/13.1 & 98.5/6.7 & 94.0/27.3 \\
        Snow & 90.8/43.2 & 95.4/21.7 & 94.4/28.4 & 96.2/19.8 & 97.2/14.0 & 94.8/25.4 \\
        Spatter & 81.5/67.1 & 86.1/60.1 & 90.7/46.0 & 93.1/36.9 & 95.3/25.3 & 89.4/47.2\\
        Speckle Noise & 87.2/52.4 & 89.2/46.7 & 93.4/31.8 & 95.0/24.9 & 96.4/18.0 & 92.2/34.8\\
        Zoom Blur & 88.7/51.1 & 91.1/44.5 & 93.0/37.0 & 94.2/32.1 & 95.5/25.6 & 92.5/38.1 \\
        \bottomrule
        \addlinespace[0.5mm]
        \textbf{Average} & 86.3/55.2 & 89.0/47.8 & 91.5/39.5 & 94.0/30.2 & 95.9/21.0 & 91.3/38.7\\
         \bottomrule
    \end{tabular}}}
\end{table}

\subsection{ASH~(ResNet-50)}

The behavior observed for ASH in Table~\ref{tab:in1k-covar-ash} is very similar to the one observed in SCALE, which uses a similar backbone.

\begin{table}[!ht]
    \centering
    \caption{Covariate shift OOD benchmark for ASH.}
    \label{tab:in1k-covar-ash}
    {\resizebox{1.0\linewidth}{!}{
    \begin{tabular}{@{\extracolsep{0pt}}lcccccc@{}}
        \toprule
        \multirow{2}{*}{\textbf{Corruption}} & \multicolumn{5}{c}{\textbf{Corruption Intensity}} & \multirow{2}{*}{\textbf{Average}}\\
        \cline{2-6}
        & \textbf{1} & \textbf{2} & \textbf{3} & \textbf{4} & \textbf{5} & \\
         \toprule
        Brightness & 64.6/86.9 & 65.5/85.3 & 67.9/82.6 & 71.9/77.5 & 77.1/69.7 & 69.4/80.4 \\
        Contrast & 79.2/68.6 & 84.7/57.4 & 92.2/35.9 & 99.0/4.9 & 100.0/0.1 & 91.0/33.4 \\
        Defocus Blur & 83.4/60.8 & 88.1/48.2 & 94.3/27.0 & 97.2/14.1 & 98.7/6.4 & 92.4/31.3 \\
        Elastic Transform & 72.0/79.4 & 79.1/69.9 & 81.8/65.5 & 87.6/52.7 & 95.2/27.6 & 83.1/59.0 \\
        Fog & 75.5/72.8 & 86.9/50.3 & 92.5/34.0 & 93.2/41.6 & 95.3/23.1 & 88.7/44.4 \\
        Frost & 75.5/72.8 & 86.9/50.3 & 92.5/34.0 & 93.2/31.6 & 95.3/23.1 & 88.7/42.4 \\
        Gaussian Blur & 75.7/75.1 & 86.4/53.4 & 92.9/33.5 & 96.4/18.5 & 99.2/4.0 & 90.1/36.9 \\
        Gaussian Noise & 70.4/79.3 & 77.8/69.9 & 88.1/48.5 & 95.7/21.7 & 99.1/3.9 & 86.2/44.7 \\
        Glass Blur & 82.0/64.8 & 89.0/46.7 & 96.3/18.8 & 97.6/12.3 & 98.7/6.5 & 92.7/29.8 \\
        Impulse Noise & 77.0/71.6 & 83.4/60.6 & 88.4/48.1 & 96.1/20.1 & 99.1/4.2 & 88.8/40.9 \\
        JPEG Compression & 71.8/80.6 & 73.8/77.7 & 75.4/75.3 & 80.8/65.3 & 87.9/48.7 & 77.9/69.5 \\
        Motion Blur & 75.8/72.7 & 82.7/59.5 & 90.8/38.4 & 96.2/18.6 & 98.0/10.0 & 88.7/39.8 \\
        Pixelate & 68.7/84.1 & 70.2/82.3 & 78.7/70.5 & 88.5/48.5 & 93.1/32.6 & 79.9/63.6 \\
        Saturate & 68.4/83.8 & 69.2/81.3 & 66.0/85.7 & 72.9/78.4 & 80.5/67.6 & 71.4/79.4 \\
        Shot Noise & 71.1/78.9 & 79.8/66.9 & 88.7/47.3 & 96.5/18.3 & 98.6/7.0 & 86.9/43.7 \\
        Snow & 75.1/75.3 & 89.4/45.0 & 86.1/52.7 & 92.1/34.9 & 95.5/21.9 & 87.6/45.9 \\
        Spatter & 64.8/87.3 & 71.5/82.1 & 79.5/72.9 & 80.5/70.6 & 86.9/58.3 & 76.6/74.2 \\
        Speckle Noise & 69.3/81.5 & 72.9/77.4 & 84.8/57.9 & 90.1/44.1 & 94.2/28.8 & 82.3/57.9 \\
        Zoom Blur & 78.8/72.6 & 83.4/64.9 & 87.0/56.2 & 89.8/48.7 & 92.0/41.1 & 86.2/56.7 \\
        \bottomrule
        \addlinespace[0.5mm]
        \textbf{Average} & 73.6/76.2 & 80.0/64.7 & 85.5/51.8 & 90.3/38.0 & 93.9/25.5 & 84.7/51.3 \\ 
         \bottomrule
    \end{tabular}}}
\end{table}

\newpage

\subsection{FDBD~(ResNet-50)}

The behavior of FDBD presented in Table~\ref{tab:in1k-covar-fdbd} is analogous to that seen in SCALE and ASH, with the three sharing the same backbone.

\begin{table}[!ht]
    \centering
    \caption{Covariate shift OOD benchmark for FDBD.}
    \label{tab:in1k-covar-fdbd}
    {\resizebox{1.0\linewidth}{!}{
    \begin{tabular}{@{\extracolsep{0pt}}lcccccc@{}}
        \toprule
        \multirow{2}{*}{\textbf{Corruption}} & \multicolumn{5}{c}{\textbf{Corruption Intensity}} & \multirow{2}{*}{\textbf{Average}}\\
        \cline{2-6}
        & \textbf{1} & \textbf{2} & \textbf{3} & \textbf{4} & \textbf{5} & \\
         \toprule
        Brightness & 62.1/89.2 & 62.7/88.5 & 64.4/86.6 & 67.8/83.3 & 72.7/77.6 & 65.9/85.0 \\
        Contrast & 76.1/75.0 & 81.7/65.5 & 89.6/46.1 & 98.2/10.0 & 99.9/0.3 & 89.1/39.4 \\
        Defocus Blur & 79.2/70.9 & 84.0/61.1 & 91.2/40.6 & 95.2/24.8 & 97.6/13.2 & 89.4/42.1 \\
        Elastic Transform & 67.2/85.1 & 74.2/76.1 & 76.9/72.7 & 83.3/61.3 & 93.1/37.4 & 79.0/66.5 \\
        Fog & 71.8/80.6 & 75.2/76.5 & 79.7/69.6 & 83.1/62.2 & 91.0/41.7 & 80.1/66.1 \\
        Frost & 71.3/79.0 & 83.6/59.4 & 90.0/43.2 & 90.9/40.3 & 93.6/30.8 & 85.9/50.5 \\
        Gaussian Blur & 72.2/81.0 & 82.4/65.0 & 90.2/45.2 & 94.7/27.6 & 98.5/7.5 & 87.6/45.3 \\
        Gaussian Noise & 71.4/80.5 & 79.2/70.2 & 88.5/49.9 & 95.5/24.1 & 99.0/4.7 & 86.7/45.9 \\
        Glass Blur & 77.8/72.5 & 85.0/57.3 & 94.0/28.0 & 95.7/20.9 & 97.3/13.6 & 90.0/38.4 \\
        Impulse Noise & 76.4/74.2 & 83.5/63.2 & 88.4/50.9 & 95.8/22.8 & 98.9/5.1 & 88.6/43.3 \\
        JPEG Compression & 70.8/83.0 & 72.9/80.3 & 74.4/78.2 & 79.2/70.6 & 85.6/56.9 & 76.6/73.8 \\
        Motion Blur & 71.2/79.6 & 78.1/68.6 & 87.7/47.9 & 94.6/25.9 & 97.0/15.2 & 85.7/47.4 \\
        Pixelate & 69.0/85.5 & 70.7/83.7 & 78.6/73.9 & 87.3/55.0 & 91.7/40.2 & 79.5/67.7 \\
        Saturate & 65.1/86.7 & 66.5/84.6 & 62.3/89.2 & 69.1/83.4 & 78.1/72.5 & 68.2/83.3 \\
        Shot Noise & 70.7/80.9 & 79.9/68.8 & 88.6/49.7 & 96.2/20.4 & 98.4/8.0 & 86.8/45.5 \\
        Snow & 72.9/78.5 & 88.0/50.4 & 84.5/57.6 & 90.8/40.8 & 94.4/27.8 & 86.1/51.0 \\
        Spatter & 61.9/89.7 & 68.6/84.6 & 77.7/75.2 & 78.7/72.6 & 85.9/60.2 & 74.6/76.5 \\
        Speckle Noise & 66.7/84.3 & 70.9/80.3 & 83.7/61.5 & 89.2/47.9 & 93.6/32.3 & 80.8/61.3 \\
        Zoom Blur & 72.9/81.1 & 77.5/76.0 & 81.5/69.6 & 84.8/63.7 & 87.6/56.7 & 80.9/69.4 \\
        \bottomrule
        \addlinespace[0.5mm]
        \textbf{Average} & 70.9/80.9 & 77.1/71.6 & 82.7/59.8 & 87.9/45.1 & 92.3/31.7 & 82.2/57.8 \\
         \bottomrule
    \end{tabular}}}
\end{table}

\subsection{DisCoPatch}

As seen in Table~\ref{tab:in1k-covar-DisCoPatch}, DisCoPatch excels at detecting every sort of corruption at each possible intensity on ImageNet-1K.

\begin{table}[!ht]
    \centering
    \caption{Covariate shift OOD benchmark for DisCoPatch-64.}
    \label{tab:in1k-covar-DisCoPatch}
    {\resizebox{1.0\linewidth}{!}{
    \begin{tabular}{@{\extracolsep{0pt}}lcccccc@{}}
        \toprule
        \multirow{2}{*}{\textbf{Corruption}} & \multicolumn{5}{c}{\textbf{Corruption Intensity}} & \multirow{2}{*}{\textbf{Average}}\\
        \cline{2-6}
        & \textbf{1} & \textbf{2} & \textbf{3} & \textbf{4} & \textbf{5} & \\
         \toprule
        Brightness & 91.4/37.2 & 91.7/34.0 & 92.7/29.5 & 93.9/24.4 & 94.8/21.4 & 92.9/29.3\\
        Contrast & 95.3/21.9 & 96.4/16.9 & 97.4/12.3 & 97.6/12.4 & 96.9/17.8 & 96.7/16.2\\
        Defocus Blur & 98.7/5.1 & 98.8/4.5 & 99.0/3.9 & 99.0/3.7 & 99.0/3.6 & 98.9/4.1\\
        Elastic Transform & 96.9/13.3 & 96.6/14.7 & 98.22/7.0 & 98.4/6.1 & 98.4/5.7 & 97.7/9.4\\
        Fog & 98.2/7.9 & 98.9/4.5 & 99.4/2.2 & 99.5/1.6 & 99.7/0.7 & 99.2/3.4\\
        Frost & 95.9/16.3 & 98.1/7.2 & 98.7/4.8 & 98.9/4.1 & 99.1/3.3 & 98.2/7.1\\
        Gaussian Blur & 98.3/7.0 & 98.8/4.4 & 99.0/3.8 & 99.0/3.6 & 99.1/3.5 & 98.8/4.5\\
        Gaussian Noise & 99.8/0.3 & 99.8/0.3 & 99.8/0.3 & 99.8/0.3 & 99.8/0.3 & 99.8/0.3\\
        Glass Blur & 98.9/4.2 & 99.2/2.8 & 99.5/1.6 & 99.5/1.4 & 99.5/1.4 & 99.3/2.3\\
        Impulse Noise & 99.8/0.5 & 99.8/0.3 & 99.8/0.3 & 99.8/0.3 & 99.8/0.3 & 99.8/0.4\\
        JPEG Compression & 83.6/55.4 & 83.1/54.0 & 82.9/53.4 & 81.2/53.6 & 78.1/56.9 & 81.8/54.7\\
        Motion Blur & 97.9/8.7 & 98.5/6.1 & 98.9/4.4 & 99.1/3.5 & 99.2/3.1 & 98.7/5.2\\
        Pixelate & 95.9/17.8 & 96.3/16.0 & 96.8/13.6 & 97.2/11.8 & 96.9/12.4 & 96.6/14.3\\
        Saturate & 95.8/18.6 & 98.5/6.2 & 93.7/29.5 & 96.4/16.0 & 97.4/11.1 & 96.3/16.3\\
        Shot Noise & 99.7/0.7 & 99.8/0.3 & 99.8/0.3 & 99.8/0.3 & 99.8/0.3 & 99.8/0.4\\
        Snow & 96.1/14.8 & 98.4/6.0 & 97.4/9.3 & 98.1/6.5 & 98.3/6.3 & 97.7/8.6\\
        Spatter & 93.0/31.3 & 95.1/19.3 & 97.3/9.9 & 94.8/17.7 & 96.6/11.2 & 95.3/17.9\\
        Speckle Noise & 99.5/2.0 & 99.6/1.4 & 99.6/1.6 & 99.5/1.7 & 99.6/1.6 & 99.5/1.6\\
        Zoom Blur & 98.4/6.7 & 98.6/5.3 & 98.8/4.5 & 98.9/4.1 & 99.0/3.7 & 98.8/4.9\\
        \bottomrule
        \addlinespace[0.5mm]
        \textbf{Average}  & 96.5/14.2 & 97.2/10.8 & 97.3/10.1 & 97.4/9.1 & 97.4/8.7 & 97.2/10.6\\ 
         \bottomrule
    \end{tabular}}}
\end{table}

\end{document}